\documentclass{article}

\usepackage[preprint]{neurips_2026}

\usepackage[utf8]{inputenc} 
\usepackage[T1]{fontenc}    
\usepackage{hyperref}       
\usepackage{url}            
\usepackage{booktabs}       
\usepackage{amsfonts}       
\usepackage{nicefrac}       
\usepackage{microtype}      
\usepackage{xcolor}         
\usepackage{graphicx}
\usepackage{amsmath}
\usepackage{multirow}
\usepackage{listings}
\usepackage{xcolor}
\usepackage{bm}
\usepackage{xurl}

\usepackage{tikz}
\usetikzlibrary{shapes.geometric, arrows.meta, positioning, calc, backgrounds, fit, matrix, shadows}

\title{FrED: External Data Influence Estimation via Domain Knowledge Graph Grounding}

\author{%
  Theodoros Aivalis\thanks{Corresponding Author. \texttt{teoaivalis@iit.demokritos.gr}} \\
  National Centre for Scientific Research ``Demokritos'', Greece \\
  University of Glasgow, UK \\
  \And
  Iraklis A. Klampanos \\
  University of Glasgow \\
  UK \\
  \AND
  Antonis Troumpoukis \\
  National Centre for Scientific Research ``Demokritos'' \\
  Greece \\
  \And
  Joemon M. Jose \\
  University of Glasgow \\
  UK \\
}

\begin{document}

\maketitle

\begin{abstract}
  The rapid deployment of generative AI has amplified the critical need for Training Data Attribution to ensure transparency and accountability. However, current parametric approaches require computationally prohibitive access to model weights, while similarity-based methods ignore deep structural context. We propose a novel probabilistic framework that operates entirely in a black-box setting. Our method fuses continuous feature similarities with discrete, domain-specific Knowledge Graphs (KGs). This approach ensures the attribution is grounded in structural reality, explicitly rewarding highly specific historical samples while preventing generic background data from dominating the results. We evaluate our framework across two distinct domains where linking outputs to data and domain context is inherently complex: abstract artistic image synthesis and high-dimensional physical weather forecasting. Extensive benchmarking demonstrates the robust efficacy of our approach. In the artistic domain, it achieves a strong Linear Datamodeling Score that exceeds standard black-box similarity baselines, while closing much of the gap to gradient-based estimators. We additionally present a cross-domain feasibility case study in environmental forecasting, where we use domain KGs to retrieve physically consistent historical analogs for regional flood forecasts, improving geographic localisation over a latent-only baseline. Operating entirely without internal model access, our approach provides an efficient, interpretable mechanism for post-hoc influence analysis and domain-grounded retrieval.
\end{abstract}

\section{Introduction}
\label{sec:introduction}

In domains of high societal impact, like medicine and Earth observation (EO), researchers want to take advantage of AI's ability to uncover hidden patterns in data to solve crucial problems, like designing novel proteins for targeted therapies or generating high-resolution forecasts grounded in regional geographical metadata. At the same time, in sensitive and creative domains, this technology has intensified the debate over explainability and intellectual property. Because most state-of-the-art models operate as ``black boxes,''\cite{koh} there is no visibility into the relationship between training samples, domain-specific knowledge and model outputs. Understanding how these systems synthesise content is critical for ensuring that generated artifacts can be reliably traced back to the training data.


For creative arts, this need for transparency is highlighted by a growing conflict between AI development and copyright law. High-profile lawsuits involving the New York Times\footnote{\url{https://harvardlawreview.org/blog/2024/04/nyt-v-openai-the-timess-about-face/}} and Disney\footnote{\url{https://digiday.com/marketing/the-disney-openai-deal-and-generative-ai-copyright-concerns/}} underscore these risks. Beyond legal disputes, over 200 artists recently signed an open letter \footnote{\url{https://www.bbc.com/news/articles/c071elp1rv1o}} urging protection against AI-driven stylistic and voice mimicking without consent, highlighting the need for royalty frameworks. In response, policy bodies have proposed new governance frameworks. The European Union’s AI Act \cite{euaiact} outlines rules for AI transparency, while the UK’s AI Strategy \cite{uk_aistrategy} and UNESCO’s guidelines \cite{unesco_ai} advocate for responsible AI. Notably, Denmark recently proposed legal amendments to grant individuals legal rights over their voice, face, and bodily likeness \footnote{\url{https://www.theguardian.com/technology/2025/jun/27/deepfakes-denmark-copyright-law-artificial-intelligence}}.

Attribution is also vital in EO for global safety. The World Meteorological Organisation's 2025 report \cite{wmo2025state} underscores that "cascading and compounding" disasters, sequential events impacting agrifood systems and migration, now affect every global region.
To process the massive data from programs like the EU's Copernicus \cite{copernicus}, meteorological agencies adopt AI-driven models such as Google GraphCast \cite{graphcast} and Huawei's Pangu-Weather \cite{pangu}. Despite their efficiency, they remain 'black boxes,' making it difficult for decision-makers to distinguish physical precedents from statistical hallucinations.
Current interpretability focus on local feature importance, yet mechanistic interpretability requires uncovering the physical relationships and interactions driving predictions. As models scale, traditional explainability tools struggle with complex data. Addressing this remains a challenge: we must move beyond post-hoc visualisations toward methods that translate mathematical outputs into historical and structural contexts, ensuring AI-driven forecasts are faithful to the fundamental physical laws.

To address these limitations, we propose FrED, a probabilistic framework for Training Data Attribution (TDA) in a strictly ``black-box'' setting. Unlike methods requiring architectural access, we estimate influence by relying solely on the relationship between model's outputs and the training data. 
Specifically, we bridge the gap between mathematical embeddings and structural understanding by grounding training data and model's outputs in domain-specific Knowledge Graphs (KGs). This allows us to combine the strengths of surface-level similarity with a networked understanding of the domain's logic. By fusing high-dimensional latent features with structured conceptual motifs, we move beyond visual correlations toward a more precise attribution method. This ensures that outputs are traced back to their origins through a combination of semantic closeness and structural relevance, providing a transparent audit trail that is faithful to both the data's content and its domain context.

Our core contributions are the following: 

\begin{enumerate}
    \item We propose FrED, a black-box influence estimator that fuses continuous latent similarity with discrete KG evidence via an asymmetric boosting.
    \item We achieve LDS performance on ArtBench that significantly outperforms black-box baselines and approaches gradient-based methods under the D-TRAK protocol.
    \item We demonstrate FrED's cross-domain utility in environmental forecasting by retrieving physically consistent historical analogs for flood forecasts.
\end{enumerate}

\paragraph{Paper Structure:} Section \ref{sec:related work} provides a review of related work in data attribution. Section \ref{sec:proposed_method} details our Bayesian framework. Section \ref{sec:experiments} presents the experimental setup and evaluation across different domains. Sections \ref{sec:conclusion} and \ref{sec:limitations} summarise the main findings and future research directions. \footnote{Source code and implementation details: \url{https://anonymous.4open.science/r/FrED-15F4/}}.

\section{Related Work}
\label{sec:related work}

We categorise current TDA research into: foundational theoretical models, state-of-the-art parametric methods, and the parameter-free observational approaches that are most relevant to our methodology.

Theoretical methods serve as the "gold standards" for the field \cite{influencesurvey}. Leave-One-Out quantifies influence by retraining models without a specific sample. While being an exact measure of data impact, it operates at an impossible computational cost for modern foundation models \cite{black2021leaveoneoutunfairness}. Similarly, game-theoretic Shapley Values provide a principled framework but suffer from exponential complexity at scale \cite{pmlrshapley}. Local Interpretable Model-agnostic Explanations (LIME) \cite{lime} approximates models locally via interpretable surrogates, but prioritises feature importance over tracing statistical lineage to training samples. To avoid full retraining, Influence Functions utilise second-order derivatives to estimate the impact of up-weighting specific samples \cite{influencesurvey}. While foundational, these methods' high costs have driven a shift toward methodologies that estimate influence without exhaustive retraining.
%
Modern parametric methods leverage internal signals, like gradients and weights, to estimate influence at scale. In Transformers, attention maps visualise specific regions or tokens \cite{attention, attention_maps}. Furthermore, Attribution by Unlearning \cite{unlearn_synthesised} approximates sample impact by directly optimising model weights to "forget" specific images and tracking resulting training loss deviations. TRAK \cite{trak} and its diffusion variant D-TRAK \cite{d-trak} enable this by using random projections to make gradient-based attribution computationally tractable. Moving beyond, the Diffusion Attribution Score (DAS) \cite{das} quantifies a sample's contribution by measuring the divergence between predicted noise distributions.
%
The final category includes methods that operate without access to model internals. At the most fundamental level, naive techniques include raw pixel similarity and high-dimensional visualization techniques like t-SNE \cite{t-SNE}, which project latent representations into low-dimensional maps to observe qualitative semantic clustering. Latent embedding similarity has emerged as the standard black-box baseline, using pre-trained encoders to measure semantic closeness. Extending these, a recent search-based influence analysis framework \cite{aivalis2025enhancing} combines textual retrieval with high-dimensional embeddings to link outputs to training samples.
Notably, Fast Data Attribution \cite{wang2025fast} distill influence scores from a computationally "slow" teacher model into a "fast" student multi-layer perceptron (MLP). Finally, the inherent difficulty of evaluation has led to benchmarking frameworks such as Attribution by Customization \cite{wang2023evaluating}, that fine-tunes generators toward a specific exemplar and creates "ground-truth" influenced images, providing a controlled environment to evaluate attribution methods.

Observational baselines identify surface-level overlap, but they lack the structural context required for semantic attribution. Recent work instead leverages Knowledge Graphs (KGs) to capture domain characteristics, enabling interpretable, concept-level comparisons. For example, KGs have been employed to trace stylistic influences by comparing the structural representations of generated and training images \cite{autographx}. Demonstrating the cross-domain adaptability, KGa are used in EO to fuse satellites and reports, enabling structural comparisons and historical retrieval of climate events \cite{extremekg}.

\section{Proposed Method}
\label{sec:proposed_method}

\subsection{A Unified Modeling Paradigm}
We establish a unified modeling paradigm for FrED, abstracting deep generative and predictive architectures as latent space transformations. While we focus on diffusion-based synthesis and weather forecasting, this formulation generalises across representation learning frameworks. Let $\mathcal{S} = \{(\mathbf{X}_j, \mathbf{T}_j)\}_{j=1}^{K}$ be the training dataset, where $\mathbf{X}_j$ is the high-dimensional observation (pixels or atmospheric tensors) and $\mathbf{T}_j$ is the associated conditioning context (text or weather states).

We decompose the model $G$ into two fundamental components \cite{bengio2017deep}:
\begin{enumerate}
    \item \textbf{The Encoder ($E$):} Maps input pairs into a compressed latent space $\mathcal{Z}$, such that $E(\mathbf{X}, \mathbf{T}) = \mathbf{z}$. This space encodes the essential statistical features of the training distribution.
    \item \textbf{The Decoder ($D$):} Projects the latent vector back into the data space to produce a reconstruction or prediction, $\mathbf{X}' = D(\mathbf{z}, \mathbf{T})$.
\end{enumerate}

Target $\mathbf{C}$ is produced by fixing the context and sampling from the latent space: $\mathbf{C} = D(\mathbf{z}_c, \mathbf{T})$. Since $\mathbf{C}$ is a product of the decoded state $\mathbf{z}_c$, we formalise attribution as tracing this state backward to identify the training samples $\mathbf{X}_j$ that most influenced the mapping $E \rightarrow D$ during optimisation.

\subsection{The Bayesian Influence Framework}
\label{sec:bayesian_framework}

\paragraph{The Ideal Influence (Counterfactual)}
To ground the concept of attribution, we define the generative process as a sampling over the latent manifold $\mathcal{Z}$. The total probability of the model generating a specific artifact $\mathbf{C}$ is expressed as:
\begin{equation}
    P(\mathbf{C}) = \int_{\mathcal{Z}} P(\mathbf{C} \mid \mathbf{z}) P(\mathbf{z}) d\mathbf{z}
\end{equation}
where $P(\mathbf{z})$ is the latent density learned from the training corpus $\mathcal{S}$ \cite{bishop2006pattern}. Ideally, the influence of a specific training sample $\mathbf{X}_j$ is defined counterfactually as 
the shift in the model's output probability for $\mathbf{C}$ that would occur if $\mathbf{X}_j$ were removed from $\mathcal{S}$. However, in a black-box setting, evaluating this shift directly is computationally prohibitive as it requires retraining for every training sample.

\paragraph{The Bayesian Proxy}
To address this, FrED approximates the counterfactual impact, approaching attribution as a posterior probability $P(\mathbf{X}_j \mid \mathbf{C})$. We assume that because the model's continuous representation is shaped by the entirety of its corpus, the generation of any target $\mathbf{C}$ can be decomposed into the discrete contributions of the training samples. For the purpose of attribution, we approximate the target as a weighted combination of the training set:
\begin{equation}
\label{eq:2}
    \mathbf{C} \approx \sum_{j=1}^{K} P(\mathbf{X}_j \mid \mathbf{C}) \mathbf{X}_j = \sum_{j=1}^{K} \alpha_j \mathbf{X}_j
\end{equation}

We note that eq. \ref{eq:2} is a ranking-oriented surrogate used to derive a practical scoring rule in black-box settings, rather than a literal generative model of diffusion or forecasting outputs.

Using Bayes' theorem, we expand the attribution weight $\alpha_j$ to identify the primary drivers of the model's density at the target coordinates:
\begin{equation}
    \alpha_j = \frac{P(\mathbf{C} \mid \mathbf{X}_j) P(\mathbf{X}_j)}{P(\mathbf{C})}
\end{equation}
For a fixed target $\mathbf{C}$, the denominator $P(\mathbf{C})$ acts as a constant normalization factor. Thus, the ranking of attribution weights is determined by the product of the likelihood and the prior:
\begin{equation}
    \alpha_j \propto P(\mathbf{C} \mid \mathbf{X}_j) P(\mathbf{X}_j)
\end{equation}

\textbf{Likelihood $P(\mathbf{C} \mid \mathbf{X}_j)$ (Latent Support):} 
This term serves as a proxy for how strongly $X_j$ supports the region of latent space from which $\mathbf{C}$ is generated.
We hypothesise that if $\mathbf{X}_j$ provides the structural instruction for a specific region of the space, any target $\mathbf{C}$ in that region is supported by $\mathbf{X}_j$. In a black-box setting, we model this via latent proximity: if the representations $\mathbf{z}_c$ and $\mathbf{z}_j$ are proximal in $\mathcal{Z}$, we assume $\mathbf{C}$ inherits its specific identity from the features of $\mathbf{X}_j$.

\textbf{Prior $P(\mathbf{X}_j)$ (Informational Scarcity):} The prior represents the informational value of $\mathbf{X}_j$ independent of the target. Moving beyond a naive uniform prior, we define this as informational scarcity. By penalising samples with dense, redundant regions in the latent space, we ensure that specialised, unique data points receive higher weight. This prevents attribution from being diluted by generic states and highlights the samples necessary to reconstruct the identity of $\mathbf{C}$.

\subsection{A Dual-Space Attribution Paradigm}
\label{sec:dual_engine}
Computing $\alpha_j$ solely via the model's latent space has limitations: deep learning architectures optimise for continuous feature similarity but remain fundamentally blind to discrete, real-world semantics. Training typically relies on sparse labels that fail to capture the rich, interconnected domain knowledge inherent in historical sample. However, for domain experts, this broader structural and contextual metadata is the primary metric for evaluating true influence and establishing trust in the model's outputs. 
To bridge machine optimization and domain context, FrED implements a retrieval-inspired fusion of two independent attribution rankings, each evaluating $\alpha_j$ within a distinct feature space.

\paragraph{Latent Space Attribution.}
Here we evaluate candidates within a high-dimensional continuous space, mirroring the encoding strategies used by the models themselves. By projecting both $\mathbf{X}_j$ and $\mathbf{C}$ into this space, we capture the continuous distributional features leveraged during the model's training phase. Within this engine, the Likelihood (Feature Similarity) measures the direct mathematical distance (vector similarity) between the training sample and the target output. Conversely, the Prior (Feature Distinctiveness) evaluates the global density of the sample within the training distribution, penalising highly redundant data to boost candidates possessing distinctive motifs.

\paragraph{Discrete Domain Ranking.}
Simultaneously, we evaluate candidates based on semantic context, which raw latent features ignore. Because real-world domain knowledge is inherently sparse, categorical, and relational, continuous embeddings fail to accurately represent it. Therefore, we can model the Domain Space as a KG, organising this discrete metadata into a verifiable networked topology. Within this KG, the Likelihood measures topological similarity between the training sample and the target artifact. The Prior evaluates structural sparsity, boosting candidates linked to highly specific, non-redundant categorical connections rather than generic, heavily crowded hubs.

This dual-mapping ensures that the final attribution reflects what the model saw (the training distribution), and also the structural reality in which that data exists (the domain context).

\subsection{Contextual Boosting for Dual-Space Fusion}
\label{sec:asymmetric_fusion}

To combine evidence from both spaces, we treat attribution as a hierarchical retrieval system. Standard rank aggregation methods (e.g. Reciprocal Rank Fusion (RRF) \cite{rrf}) treat data sources as equal contributors. However, because generative and predictive models are optimised primarily on continuous latent features, treating discrete metadata with equal contribution can degrade the primary signal.
Moreover, this approach provides robustness against incomplete or non-representative domain knowledge. If a sample lacks connections in the KG, an equal-weighting fusion would penalise it. By structuring the domain evidence as an asymmetric boost, system uses foundational latent score when metadata is absent, ensuring that attributions are never discarded due to missing graph topology.

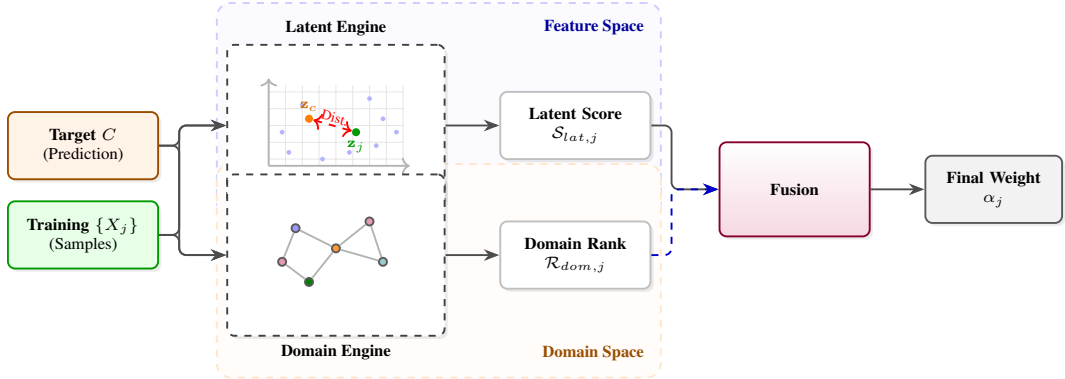
\begin{figure}[htbp]
\centering
\resizebox{\textwidth}{!}{%
\begin{tikzpicture}[
    basebox/.style={draw, rectangle, align=center, rounded corners=3pt, font=\scriptsize, drop shadow={opacity=0.15, shadow xshift=0.3ex, shadow yshift=-0.3ex}},
    inputbox/.style={basebox, fill=gray!10, draw=gray!70!black, thick, text width=2.0cm, minimum height=1.0cm},
    enginebox/.style={basebox, fill=white, dashed, draw=gray!50!black, thick, text width=3.0cm, minimum height=2.4cm},
    scorebox/.style={basebox, fill=white, draw=gray!50, thick, text width=2.0cm, minimum height=1.0cm},
    fusionbox/.style={basebox, top color=white, bottom color=purple!15, draw=purple!70!black, thick, minimum height=1.4cm, text width=2cm},
    arrow/.style={-{Stealth[scale=1.0]}, thick, draw=black!70, rounded corners=3pt},
    dashed_arrow/.style={arrow, dashed, draw=blue!80!black}
]

\node (target) [inputbox, fill=orange!10, draw=orange!60!black] {\textbf{Target $C$}\\ (Prediction)};
\node (corpus) [inputbox, below=0.3cm of target, fill=green!10, draw=green!60!black] {\textbf{Training $\{X_j\}$}\\ (Samples)};


\node (latent_engine) [enginebox, right=1.0cm of target, yshift=0.3cm] {};
\node (latent_title) at (latent_engine.north) [above, font=\scriptsize\bfseries] {Latent Engine};

\begin{scope}[shift={(latent_engine.center)}]
    \draw[step=0.25cm, gray!20, very thin] (-1.0,-0.6) grid (1.0,0.6);
    \draw[->, gray!60, thick] (-1.0, -0.6) -- (1.1, -0.6);
    \draw[->, gray!60, thick] (-1.0, -0.6) -- (-1.0, 0.7);
    
    \foreach \x/\y in {-0.7/-0.4, -0.8/-0.1, -0.5/0.3, 0.2/-0.4, 0.6/-0.3, 0.7/0.1, 0.9/-0.1, 0.5/0.4, -0.2/-0.5} {
        \fill[blue!30] (\x,\y) circle (1.0pt);
    }
    
    \node (zc) at (-0.4, 0.1) [circle, fill=orange, inner sep=1.2pt, label={[font=\tiny, text=orange!80!black, label distance=-2pt]above:$\mathbf{z}_c$}] {};
    \node (zj) at (0.3, -0.1) [circle, fill=green!60!black, inner sep=1.2pt, label={[font=\tiny, text=green!60!black, label distance=-2pt]below:$\mathbf{z}_j$}] {};
    
    \draw[<->, dashed, red, thick] (zc) -- (zj) node[midway, sloped, above, font=\tiny, inner sep=1pt] {Dist.};
\end{scope}

\node (domain_engine) [enginebox, right=1.0cm of corpus, yshift=-0.3cm] {};
\node (domain_title) at (domain_engine.south) [below, font=\scriptsize\bfseries] {Domain Engine};

\begin{scope}[shift={(domain_engine.center)}]
    \tikzstyle{kg_node}=[circle, draw=gray!70!black, thick, inner sep=1.2pt]
    
    \node (kg_center) at (0, 0.1) [kg_node, fill=orange!80] {};
    \node (kg_1) at (-0.6, 0.4) [kg_node, fill=blue!40] {};
    \node (kg_2) at (0.5, 0.5)  [kg_node, fill=purple!40] {};
    \node (kg_3) at (0.7, -0.1) [kg_node, fill=teal!40] {};
    \node (kg_4) at (-0.4, -0.4)[kg_node, fill=green!50!black] {};
    \node (kg_5) at (-0.8, -0.1)[kg_node, fill=purple!40] {};
    
    \draw[gray!60, thick] (kg_center) -- (kg_1);
    \draw[gray!60, thick] (kg_center) -- (kg_2);
    \draw[gray!60, thick] (kg_center) -- (kg_3);
    \draw[gray!60, thick] (kg_center) -- (kg_4);
    
    \draw[gray!60, thick] (kg_1) -- (kg_5);
    \draw[gray!60, thick] (kg_4) -- (kg_5);
    \draw[gray!60, thick] (kg_2) -- (kg_3);
\end{scope}

\node (latent_score) [scorebox, right=0.8cm of latent_engine] {\textbf{Latent Score}\\$\mathcal{S}_{lat, j}$};
\node (domain_rank)  [scorebox, right=0.8cm of domain_engine] {\textbf{Domain Rank}\\$\mathcal{R}_{dom, j}$};

\node (fusion) [fusionbox, right=1.0cm of latent_score, yshift=-0.95cm] {\textbf{Fusion}};

\node (output) [inputbox, right=0.8cm of fusion, text width=1.8cm] {\textbf{Final Weight}\\$\alpha_j$};

\draw [arrow] (target.east) -- ++(0.3,0) |- (latent_engine.west);
\draw [arrow] (target.east) -- ++(0.3,0) |- (domain_engine.west);
\draw [arrow] (corpus.east) -- ++(0.3,0) |- (latent_engine.west);
\draw [arrow] (corpus.east) -- ++(0.3,0) |- (domain_engine.west);

\draw [arrow] (latent_engine.east) -- (latent_score.west);
\draw [arrow] (domain_engine.east) -- (domain_rank.west);

\draw [arrow] (latent_score.east) -- ++(0.3,0) |- (fusion.west);
\draw [dashed_arrow] (domain_rank.east) -- ++(0.3,0) |- (fusion.west);

\draw [arrow] (fusion.east) -- (output.west);

\begin{scope}[on background layer]
    \node (latent_group) [draw=blue!20, fill=blue!2, rounded corners=6pt, inner sep=4pt, dashed, thick, fit=(latent_title) (latent_engine) (latent_score)] {};
    \node at (latent_group.north east) [anchor=north east, font=\scriptsize\bfseries\color{blue!60!black}, xshift=-4pt, yshift=-4pt] {Feature Space};

    \node (domain_group) [draw=orange!20, fill=orange!2, rounded corners=6pt, inner sep=4pt, dashed, thick, fit=(domain_title) (domain_engine) (domain_rank)] {};
    \node at (domain_group.south east) [anchor=south east, font=\scriptsize\bfseries\color{orange!60!black}, xshift=-4pt, yshift=4pt] {Domain Space};
\end{scope}

\end{tikzpicture}%
} 
\caption{\textbf{FrED}'s overview showing the fusion between latent and domain spaces. A hierarchical mechanism applies the discrete domain rank as a contextual boost to the continuous latent score.}
\label{fig:architecture}
\end{figure}

Inspired from retrieval systems, we use the Latent score as primary evidence and the Domain rank as a contextual guide. As Figure \ref{fig:architecture} shows, the final weight $\alpha_j$ for a training sample $\mathbf{X}_j$ is calculated as:

\begin{equation}
    \alpha_j = \mathcal{S}_{lat, j} \times \left( 1.0 + \text{Boost}(\mathcal{R}_{dom, j}) \right)
\end{equation}

where the boosting term uses the rank-decay principles of RRF, applied to the Domain Engine:

\begin{equation}
    \text{Boost}(\mathcal{R}_{dom, j}) = \frac{W}{k + \mathcal{R}_{dom, j}}
\end{equation}

Here, $\mathcal{S}_{lat, j}$ is the Latent Engine's base probability (combining both the latent similarity and the distinctiveness prior). $\mathcal{R}_{dom, j}$ is the discrete integer rank of the sample in the Domain Engine. $W$ defines the maximum scaling weight, and $k$ is a constant to control how fast the boost drops off. 

\section{Experiments}
\label{sec:experiments}
%


We evaluate \textbf{FrED} across two domains with distinct attribution challenges and evaluation paradigms. In Art Generation (Sec.~\ref{sec:art_domain}), we benchmark against parametric methods on ArtBench to trace stylistic lineages. In Environmental Forecasting (Sec.~\ref{sec:env_domain}), prohibitive retraining costs for global models prevent standard counterfactual evaluation. We instead present a feasibility study for analog retrieval, demonstrating our versatility in aligning high-dimensional physical states with structural knowledge.

\subsection{The Artistic Domain}
\label{sec:art_domain}

\subsubsection{Datasets and Domain Modeling}
\label{sec:art_datasets}

We utilise the ArtBench-10 dataset \cite{artbench}, a standard benchmark for generative modeling consisting of class-balanced artworks categorised into ten artistic styles. While the raw dataset provides the visual modality and basic labels, it lacks the deep historical context necessary for structural attribution. 
To bridge this gap and establish our domain knowledge, we curate a comprehensive metadata registry for every painting. Using an automated extraction pipeline, we query WikiArt, Wikipedia, and DBpedia to collect historical and semantic information across two primary levels:
\begin{enumerate}
    \item \textbf{Painting-Level Metadata:} Specific categorical details for each artwork, mapping the image to its explicit genre, style, artistic movement and descriptive tags.
    \item \textbf{Artist-Level Metadata:} Rich biographical and historical data capturing the artist's active years, geographical associations, academic institutions, and complex interpersonal lineages (e.g., who influenced them, who they taught, and their professional collaborators).
\end{enumerate}

By aggregating these heterogeneous data, we construct the Domain KG ($\mathcal{G}_{D}$), that encodes the historical interconnections between the training samples. The overall domain schema is detailed in Appendix \ref{app:art_kg}. Figure \ref{fig:art_kg_example} presents a visual example paired with its corresponding extracted subgraph.

\begin{figure}[htbp]
    \centering
    \begin{minipage}{0.40\linewidth}
        \centering
        \includegraphics[width=\linewidth]{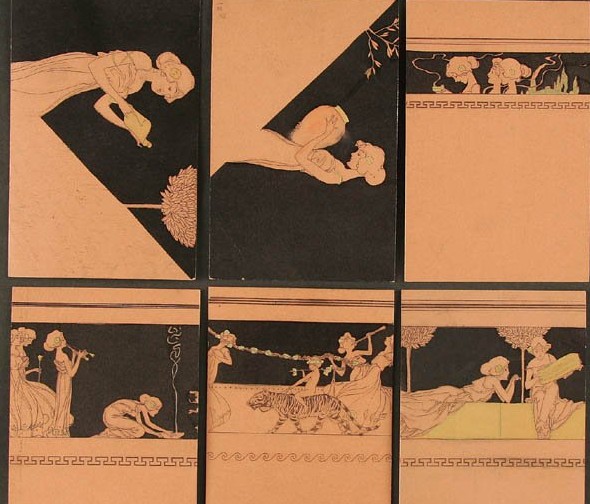}
        \centerline{(a) Visual Exemplar: Scenes of Ancient Greece}
    \end{minipage}\hfill
    \begin{minipage}{0.32\linewidth}
        \centering
        \includegraphics[width=\linewidth]{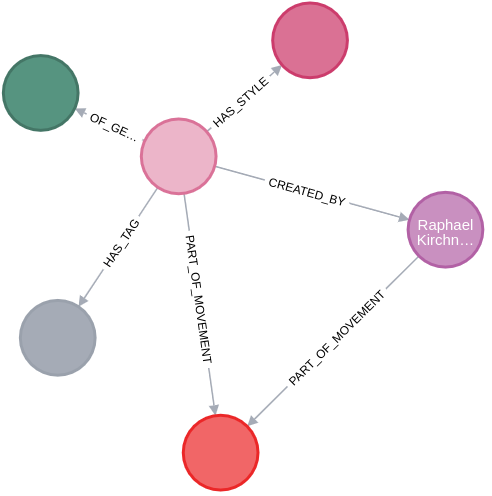}
        \centerline{(b) Extracted KG for the Exemplar}
    \end{minipage}
    \caption{Artistic exemplar: (a) visual data and (b) its extracted KG, demonstrating the mapping of visual features to topological nodes.}
    \label{fig:art_kg_example}
\end{figure}

\paragraph{Domain Alignment via Contrastive Learning.}
A challenge in structural attribution is evaluating generated images ($\mathbf{C}$) against the domain KG. These images lack the metadata required to position them within our domain. Similarly, many training samples ($\mathbf{X}_j$) have missing records. 
To solve this, we create a mapping that translates visual features into graph coordinates. We curate a representative subset of training images with rich metadata, using them as ground-truth anchors. We then adopt a contrastive learning paradigm inspired by CLIP \cite{clip_contrastive}, but instead of aligning images with text, we optimise the alignment between visual embeddings and their graph representations.
For methodological consistency with our latent engine, we utilise a pre-trained ViT-g-14 \cite{openclip} and apply node2vec \cite{node2vec} in the graph modality. By optimising a symmetric InfoNCE loss, we force visual and structural features to align in a shared representation space. Once this space is established, we project any target image in the domain space, enabling us to compute the structural ranking ($\mathcal{R}_{dom, j}$), providing the necessary foundation for our method. Appendix \ref{app:contrastive_learning} details data selected, the encoders and training procedure.

\subsubsection{Generative Setup and Evaluation Protocol}
\label{sec:art_generative_setup}
To ensure direct comparison with state-of-the-art parametric baselines, we follow the experimental protocol established by D-TRAK \cite{d-trak} for ArtBench. We use the same generative pipeline, with identical model architectures, hyperparameter configurations, and training schedules. We represent our attribution engines using a pre-trained ViT-g-14 for latent features and Node2Vec for structural domain topology.
Crucially, we evaluate our framework using the exact same set of 1,000 generated target artifacts ($\mathbf{C}$) synthesised for the baseline's evaluation. By maintaining consistency across training subsets and outputs, we isolate attribution performance as the sole independent variable, ensuring gains reflect our methodology rather than variations in the underlying generative model.

%

\paragraph{Generative Setup: Stable Diffusion and LoRA.}
Following the established benchmark, we fine-tune a $256 \times 256$ Stable Diffusion model \cite{rombach2022high} using LoRA \cite{hu2022lora} ($r=128$, 25.5M parameters). Training is class-conditional, utilising prompts such as ``a \{class\} painting''. To ensure comparison with baselines, we replicate the D-TRAK training and synthesis pipeline. Specifics regarding optimisation schedules, data augmentation, and DDIM solver settings are detailed in Appendix \ref{app:hyperparameters}.

\paragraph{Evaluation Metric: The Linear Datamodeling Score.}
Evaluating attribution requires the predicted scores accurately reflect actual causal dynamics during the generative process. We validate FrED via the Linear Datamodeling Score (LDS), introduced by TRAK \cite{park2023trak} and adapted for diffusion models by D-TRAK \cite{d-trak} and DAS \cite{das}.
The premise of LDS is that an effective method must act as a reliable counterfactual predictor. If a framework identifies the important training points for a generation, modifying them should lead to a predictable change in the model's output. LDS calculates the Spearman rank correlation ($\rho$) between two distinct variables across an ensemble of subsets:
\begin{enumerate}
    \item \textbf{Actual Behavior:} The ground-truth change in model output when the model is trained on random subsets of the training data. Following the benchmark, this comprises an ensemble of models trained across 64 distinct subsets, each containing 50\% of the original data.
    \item \textbf{Attribution Prediction:} The additive proxy, representing the sum of the predicted attribution scores assigned to the specific training samples included in those random subsets.
\end{enumerate}
A high LDS ($\rho$) indicates a strong relationship between our framework's predicted scores and the actual behavior of the generative model under counterfactual conditions. The final reported LDS is the average score across all evaluated target artifacts. For completeness, we provide the extended mathematical formulation of the LDS ensemble procedure in Appendix \ref{app:lds_math}.

\subsubsection{Quantitative Benchmarking and Results}
\label{sec:art_results}

Table \ref{tab:artbench_side_by_side_complete} presents the LDS evaluation across both Validation (real images) and Generation (synthetic images) sets for the ArtBench-2 and ArtBench-5 splits. We categorise baselines in two groups based on their level of model access: Non-Parametric methods, and Parametric methods that require internal access. Detailed implementation specifics for these baseline methods are provided in Appendix \ref{app:baselines}.

\begin{table*}[ht]
\centering
\caption{LDS results (\%) on ArtBench-2 and ArtBench-5. \textbf{FrED} significantly outperforms black-box baselines and closes the gap to gradient-based estimators without requiring internal model access.}
\label{tab:artbench_side_by_side_complete}
\resizebox{\textwidth}{!}{%
\begin{tabular}{lcccccccc}
\toprule
 & \multicolumn{4}{c}{\textbf{ArtBench-2}} & \multicolumn{4}{c}{\textbf{ArtBench-5}} \\
\cmidrule(lr){2-5} \cmidrule(lr){6-9}
 & \multicolumn{2}{c}{Validation} & \multicolumn{2}{c}{Generation} & \multicolumn{2}{c}{Validation} & \multicolumn{2}{c}{Generation} \\
\cmidrule(lr){2-3} \cmidrule(lr){4-5} \cmidrule(lr){6-7} \cmidrule(lr){8-9}
\textbf{Method} & \textbf{10} & \textbf{100} & \textbf{10} & \textbf{100} & \textbf{10} & \textbf{100} & \textbf{10} & \textbf{100} \\
\midrule
\multicolumn{9}{l}{\textit{\textbf{Non-Parametric methods}}} \\
{Raw pixel (dot prod.)} & \multicolumn{2}{c}{$2.4 \pm 1.1$} & \multicolumn{2}{c}{$2.5 \pm 1.5$} & \multicolumn{2}{c}{$1.84 \pm 0.42$} & \multicolumn{2}{c}{$2.77 \pm 0.80$} \\
{Raw pixel (cosine)} & \multicolumn{2}{c}{$2.5 \pm 1.2$} & \multicolumn{2}{c}{$2.6 \pm 1.5$} & \multicolumn{2}{c}{$1.97 \pm 0.41$} & \multicolumn{2}{c}{$3.22 \pm 0.78$} \\
{CLIP similarity (dot prod.)} & \multicolumn{2}{c}{$7.3 \pm 1.5$} & \multicolumn{2}{c}{$5.3 \pm 2.4$} & \multicolumn{2}{c}{$5.29 \pm 0.45$} & \multicolumn{2}{c}{$4.47 \pm 1.09$} \\
{CLIP similarity (cosine)} & \multicolumn{2}{c}{$8.9 \pm 1.3$} & \multicolumn{2}{c}{$8.8 \pm 2.2$} & \multicolumn{2}{c}{$6.57 \pm 0.44$} & \multicolumn{2}{c}{$6.63 \pm 1.14$} \\
\textbf{FrED (Ours)} & \multicolumn{2}{c}{\bm{$29.1 \pm 0.65$}} & \multicolumn{2}{c}{\bm{$21.3 \pm 1.65$}} & \multicolumn{2}{c}{\bm{$28.9 \pm 0.35$}} & \multicolumn{2}{c}{\bm{$18.4 \pm 1.05$}} \\
\midrule
\multicolumn{9}{l}{\textit{\textbf{Parametric methods}}} \\
Gradient (dot prod.) \cite{charpiat2019input} & $7.68 \pm 0.43$ & $16.00 \pm 0.51$ & $4.07 \pm 1.07$ & $10.23 \pm 1.08$ & $4.77 \pm 0.36$ & $10.02 \pm 0.45$ & $3.89 \pm 0.88$ & $8.17 \pm 1.02$ \\
Gradient (cosine) \cite{charpiat2019input} & $7.72 \pm 0.42$ & $16.04 \pm 0.49$ & $4.50 \pm 0.97$ & $10.71 \pm 1.07$ & $4.96 \pm 0.35$ & $9.85 \pm 0.44$ & $4.14 \pm 0.86$ & $8.18 \pm 1.01$ \\
{TracInCP} \cite{pruthi2020} & $9.69 \pm 0.49$ & $17.83 \pm 0.58$ & $6.36 \pm 0.93$ & $13.85 \pm 1.01$ & $5.33 \pm 0.37$ & $10.87 \pm 0.47$ & $4.34 \pm 0.84$ & $9.02 \pm 1.04$ \\
GAS \cite{hammoudeh2022identifying} & $9.65 \pm 0.46$ & $18.04 \pm 0.62$ & $6.74 \pm 0.82$ & $14.27 \pm 0.97$ & $5.52 \pm 0.38$ & $10.71 \pm 0.48$ & $4.48 \pm 0.83$ & $9.13 \pm 1.01$ \\
Journey TRAK \cite{georgiev2023journey} & -- & -- & $5.96 \pm 0.97$ & $11.41 \pm 1.02$ & -- & -- & $7.59 \pm 0.78$ & $13.31 \pm 0.68$ \\
Relative IF \cite{barshan2020relatif} & $12.22 \pm 0.43$ & $27.25 \pm 0.34$ & $7.62 \pm 0.57$ & $19.78 \pm 0.69$ & $9.77 \pm 0.34$ & $20.97 \pm 0.41$ & $8.89 \pm 0.59$ & $19.56 \pm 0.62$ \\
Renorm. IF \cite{hammoudeh2022identifying} & $11.90 \pm 0.43$ & $26.49 \pm 0.34$ & $7.83 \pm 0.64$ & $19.86 \pm 0.71$ & $9.57 \pm 0.32$ & $20.72 \pm 0.40$ & $8.97 \pm 0.58$ & $19.38 \pm 0.66$ \\
TRAK \cite{trak} & $12.26 \pm 0.42$ & $27.28 \pm 0.34$ & $7.78 \pm 0.59$ & $20.02 \pm 0.69$ & $9.79 \pm 0.33$ & $21.03 \pm 0.42$ & $8.79 \pm 0.59$ & $19.54 \pm 0.61$ \\
D-TRAK \cite{d-trak} & $27.61 \pm 0.49$ & $32.38 \pm 0.41$ & $24.16 \pm 0.67$ & $26.53 \pm 0.64$ & $22.84 \pm 0.37$ & $27.46 \pm 0.37$ & $21.56 \pm 0.71$ & $23.85 \pm 0.74$ \\
DAS \cite{das} & \bm{$37.96 \pm 0.64$} & \bm{$40.77 \pm 0.47$} & \bm{$30.81 \pm 0.31$} & \bm{$32.31 \pm 0.42$} & \bm{$35.33 \pm 0.49$} & \bm{$37.67 \pm 0.68$} & \bm{$31.74 \pm 0.75$} & \bm{$32.77 \pm 0.53$} \\

\bottomrule
\end{tabular}%
}
\end{table*}

\paragraph{Encoder Choice and Accuracy Boost.} 
The high accuracy of FrED, as evidenced in Table \ref{tab:artbench_side_by_side_complete}, is based on the selection of ViT-g-14 \cite{openclip}. As demonstrated in our ablation studies (Table \ref{tab:encoder_ablation_combined}), this model yields superior predictive faithfulness, justified by the scale and diversity of the LAION-2B distribution. This high-diversity set allows the encoder to develop a generalised feature space that captures the fine-grained stylistic characteristics necessary for art attribution. Furthermore, we found that applying a non-linear scaling exponent, specifically $p(c|x_j)^{10}$, is critical for mathematical sharpening. Since influence is often concentrated in a sparse subset of training data, this mechanism amplifies causal signals while driving the attribution of non-influential points toward zero.

\paragraph{Engine Validation and Fusion.} 
Our experiments reveal that both the Latent and Domain spaces possess significant individual attribution power (Table \ref{tab:ablation_study}). 
However, the comparison of fusion techniques highlights the limitations of linear combinations (Hybrid 50/50), which often introduce categorical noise to the visual signals. Our Boost Technique treats the Domain Space as topological evidence rather than a parallel signal. By using the KG strictly to re-rank candidates from the Latent Space, we ensure the final attribution is both visually acceptable and historically consistent. This asymmetric synthesis allows our framework to bridge the performance gap with gradient-intensive parametric methods while maintaining the efficiency and accessibility of a black-box system.

\subsection{Case Study: Domain-Grounded Retrieval of Historical Analogs for Flood Forecasts}
\label{sec:env_domain}


Moving beyond art, we apply FrED scoring pipeline strictly for domain-grounded analog retrieval. Since counterfactual evaluation is currently impractical for weather models, our metrics assess regional fidelity and physical consistency, not training-data attribution, and do not imply evidence of causality, memorisation, or training-set membership.

\subsubsection{Data Collection and Environmental Mapping}
\label{sec:eo_datasets}
 We combine different data in an Environmental KG ($\mathcal{G}_{E}$), connecting weather patterns to species and ground conditions on the Earth.
We utilise 1,372 flood events from \cite{extremekg}. Each record includes a unique ID, exact dates, and impact data (e.g., damage and displacement). Precise GPS coordinates link meteorological records to environmental proxies. To mitigate observation bias, we treat this domain knowledge strictly as a secondary re-ranking signal across two levels:

\begin{enumerate}
    \item \textbf{The Nature Layer (Biological):} Utilising iNaturalist \cite{inaturalist2026} sightings within a 10km radius, we capture regional ecological signatures to integrate biodiversity patterns into our KG. This enables cross-regional comparisons based on ecological habitats. 
    \item \textbf{The Land Cover Layer (Geographic):} Using ESA LandCover satellites \cite{zanaga2022esa}, we analyse surface composition in a 10km radius. Integrating these geographic signatures, such as urban, agricultural, or water-absorbing zones, allows the framework to evaluate how ground conditions influence forecast severity and link regions with similar physical vulnerabilities.
\end{enumerate}

    %


Integrating these data into a unified graph architecture creates a dense network connecting meteorological events to their surroundings (details in Appendix \ref{app:environment_kg}). Figure \ref{fig:env_kg_example} demonstrates this by pairing a specific flood event with its extracted environmental subgraph.

Unlike the art study’s contrastive inference, here we utilise Direct Geospatial Matching. Since forecasts provide precise coordinates and timestamps, we query the exact environmental state from the KG. This links current forecasts to historical analogs sharing identical environmental constraints.

\begin{figure}[htbp]
    \centering
    \begin{minipage}{0.50\linewidth}
        \centering
        \includegraphics[width=\linewidth]{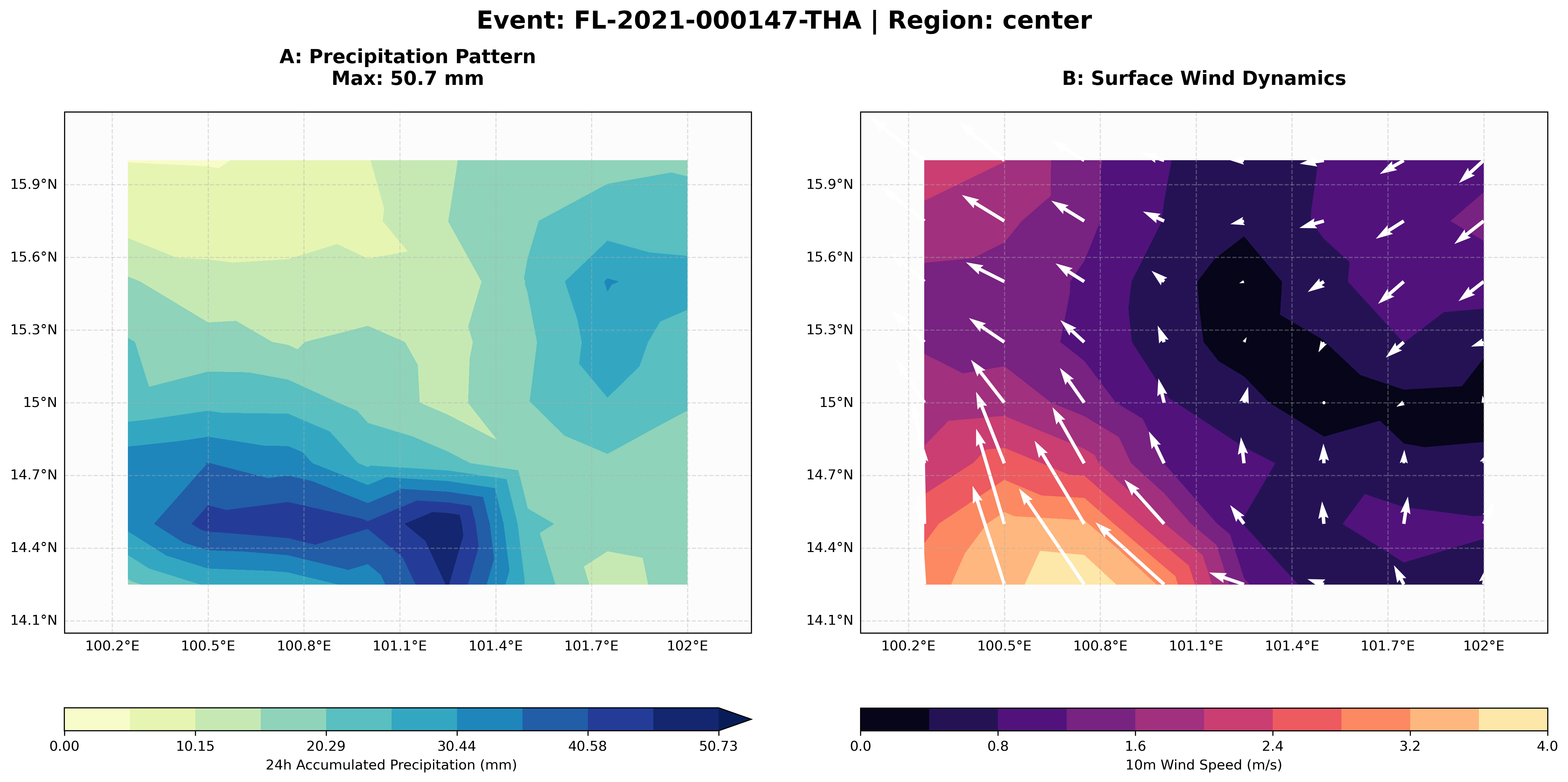}
        \centerline{(a) Meteorological Exemplar: FL-2021-000147}
    \end{minipage}\hfill
    \begin{minipage}{0.40\linewidth}
        \centering
        \includegraphics[width=\linewidth]{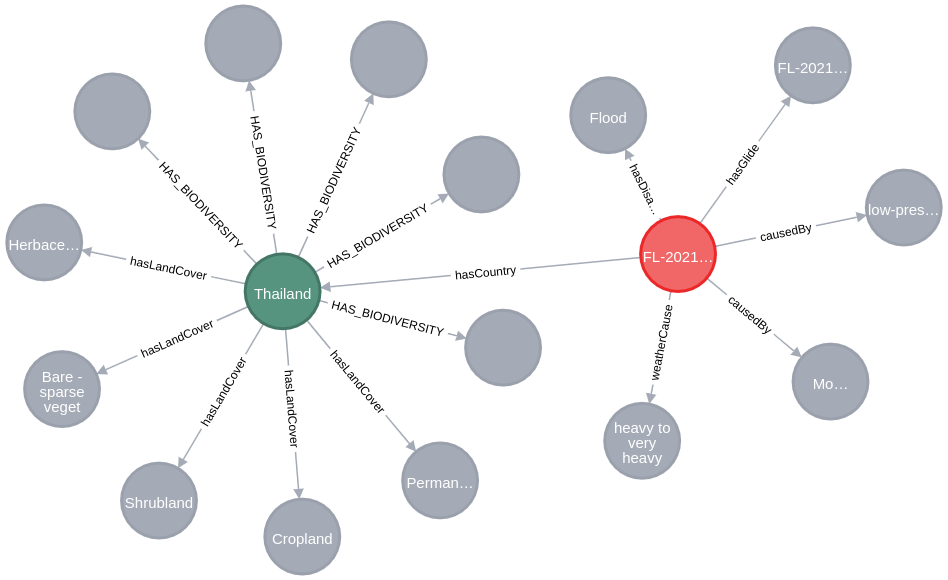}
        \centerline{(b) Extracted KG for the Event}
    \end{minipage}
    \caption{Environmental exemplar: (a) meteorological data and (b) its extracted KG, demonstrating the mapping of atmospheric variables to structural nodes.}
    \label{fig:env_kg_example}
\end{figure}
\vspace{-0.3cm}

\subsubsection{Predictive Setup and Evaluation Targets}
\label{sec:env_predictive_setup}

We prioritise regional meteorological dynamics over global predictions. While foundation models process data globally, effective analog retrieval grounding requires higher resolution to isolate the local extremes and geographic traits. Full-globe embeddings would encode data from unrelated regions, weakening local influence and preventing experts from tracing clear contextual links. Localised filtering instead ensures the framework captures the unique atmospheric signatures of each flood.

\vspace{-0.2cm}
\paragraph{Predictive Baseline: Pangu-Weather and WeatherBench 2.}
We utilise Pangu-Weather \cite{pangu} via WeatherBench2 \cite{rasp2023weatherbench} for 237 flood events (2018--2022), defining target ($\mathbf{C}$) as the 24-hour forecast at peak intensity.
Unlike the 2D pixel grids of the artistic domain, retrieving historical precedents in weather forecasting requires evaluating complex atmospheric tensors.
While Pangu operates globally at $0.25^{\circ}$ resolution, we extract regional patches centered on event coordinates to capture local signatures. Within these patches, we subset the tensor to isolate primary flood drivers, including surface variables and upper-air profiles across four pressure levels. 
For representation, we train a 3D Autoencoder for latent features and employ Node2Vec for structural graph topology. Comprehensive details on architecture, variable selection, and training protocols are provided in Appendix \ref{app:weather_setup}.

\paragraph{Evaluation Protocol: Regional Fidelity and Ecological Consistency.} 
Unlike the artistic domain, where LDS is the gold standard, attribution for global weather foundation models remains unexplored. We evaluate our environmental case-study as domain-grounded analog retrieval, not validated training-data influence.
Consequently, we assess the quality of our work through:
\begin{enumerate}
    \item \textbf{Geographic Precision:} We quantify the percentage of top-$K$ historical analogs that share the target forecast's country. This quantifies the model's ability to anchor continuous atmospheric states to specific geographic locations.
    \item \textbf{Ecological Consistency:} For cross-country matches, we evaluate similarity using Landcover Overlap and Species Category Similarity. This validates whether we retrieve physically consistent analogs when exact geographic matches are unavailable.
\end{enumerate}

\subsubsection{Results and Physical Consistency}
\label{sec:env_results}

Table \ref{tab:rmr_results} shows that our method significantly improves regional localisation, reaching a Geographic Precision of $65.7\%$ compared to $50.9\%$ for the baseline, demonstrating that the KG situates abstract atmospheric data within correct physical locations. 
This shift moves from standard retrieval based on numerical similarity,
toward a framework grounded in physical consistency. 
While precision decreases as $K$ increases due to limited regional history, FrED consistently outperforms the baseline.

\begin{table}[ht]
\centering
\caption{\textbf{FrED} improves geographic precision by anchoring atmospheric states to regional contexts. Columns report Geographic Precision (\%) at top-$K$ ranks ($K$ = 10, 20, 50, 100).}
\label{tab:rmr_results}
\small
\begin{tabular}{@{}lcccc@{}}
\toprule
\textbf{Retrieval Method} & \textbf{Prec@10} & \textbf{Prec@20} & \textbf{Prec@50} & \textbf{Prec@100} \\ \midrule
Latent Baseline & 50.9\% & 35.6\% & 15.2\% & 6.8\% \\
\textbf{FrED (Ours)} & \textbf{65.7\%} & \textbf{47.8\%} & \textbf{22.5\%} & \textbf{12.0\%} \\ \bottomrule
\end{tabular}
\end{table}


Retrievals outside the target country maintain high ecological consistency, averaging \textbf{42.1\% Landcover Overlap} and \textbf{46.2\% Species Similarity}. A representative example is the linking of West African savanna profiles in Gambia--Burkina Faso. Similarly, recurring substitutions like South Sudan--Mali, Central African Republic--Sudan, and Paraguay--Chile demonstrate that FrED identifies physically consistent analogs by matching structural signatures across disparate geographic regions.


\vspace{-0.2cm}
\section{Conclusion}
\label{sec:conclusion}

In this paper, we introduced FrED, a probabilistic framework for TDA in black-box settings. Unlike parametric methods, we estimate influence through the hierarchical fusion of continuous latent features and domain-specific KGs. Moving the attribution mechanism outside the model's architecture, we provide a scalable solution for auditing foundation models that are inaccessible due to computational constraints.
We demonstrated the versatility of our methodology across two domains. In the artistic domain, benchmarking on ArtBench revealed that we achieve an LDS competitive with several parametric estimators, while requiring zero internal access.
In the environmental domain, we demonstrate feasibility via domain-grounded analog retrieval; establishing counterfactual causal training influence for such models remains future work. By prioritising regional characteristics, we retrieve physically relevant historical analogs for high-dimensional forecasts. 
%
%
Overall, our findings suggest that grounding latent similarity with domain context is a powerful mechanism for producing interpretable, domain-consistent influence hypotheses and historical analogs in black-box settings.
%

\vspace{-0.2cm}
\section{Limitations and Future Directions}
\label{sec:limitations}
Key limitations shape our future research directions. The framework's reliance on domain KGs may present an ``expert bottleneck'' in some fields, suggesting future collaboration to formalise domain knowledge. Additionally, as KGs provide static snapshots, we aim to implement real-time graph updates. Our evaluation is further constrained by the computational intensity of the LDS metric, suggesting that one might develop more efficient, domain-specific benchmarks or try rank-targeted fusion strategies to improve accuracy without full model retraining. Where original training sets are inaccessible, we can utilise representative KG samples or approximate distributions via targeted web-scraping, similarly to \cite{aivalis2025enhancing}. Finally, we will investigate applications in medicine and biology, where tracing sequences back to clinical datasets offers a vital path toward safe, AI-driven discovery.

{
\small
\bibliographystyle{unsrt}
\bibliography{references}
}

\appendix


\section{Artistic KG: Schema and Construction}
\label{app:art_kg}

To establish the topological prior for the artistic domain, we constructed a comprehensive Domain KG ($\mathcal{G}_D$). Rather than relying solely on the sparse labels provided by standard generative datasets, we followed an automated data extraction pipeline to build a deeply interconnected historical network.

To establish the topological prior for the artistic domain, we constructed a comprehensive Domain KG ($\mathcal{G}_D$). By moving beyond the simple categorical labels of the Artbench dataset, we implemented an automated retrieval workflow to build a deeply interconnected historical network.

\subsection{Data Collection and Sequential Enrichment}
The data ingestion process followed a multi-stage enrichment strategy designed to transform a flat vision dataset into a relational network. 

\begin{enumerate}
    \item \textbf{Primary Ingestion:} We initiated the pipeline using painting identifiers and artist names provided by ArtBench-10.
    \item \textbf{Metadata Expansion (WikiArt):} Using these identifiers as primary keys, we queried the WikiArt API to systematically retrieve all available technical and categorical metadata for each artwork. We then performed a secondary expansion to ingest granular biographical profiles for the associated artists, including active years and thematic tags.
    \item \textbf{Relational Resolution (Wikipedia \& DBpedia):} To capture all the possible historical context, we utilised the Wikipedia URLs embedded within the WikiArt records as bridge identifiers to aggregate data from both Wikipedia and DBpedia. From Wikipedia, we extracted available biographical summaries and historical descriptions of each artist. Simultaneously, we leveraged the structured nature of DBpedia by executing targeted SPARQL queries to ingest specific relational attributes. By combining these sources, we were able to map complex interpersonal networks, effectively recovering the structural context that is inherently missing from the raw vision data.
\end{enumerate}

\begin{figure}[htbp]
    \centering
    \begin{minipage}{0.35\linewidth}
        \centering
        \includegraphics[width=\linewidth]{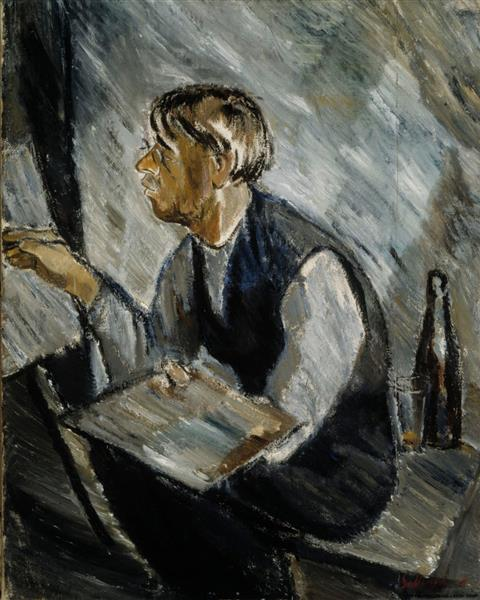}
        \centerline{(a) Artwork: Ruokokoski Maalaa}
    \end{minipage}\hfill
    \begin{minipage}{0.50\linewidth}
        \centering
        \includegraphics[width=\linewidth]{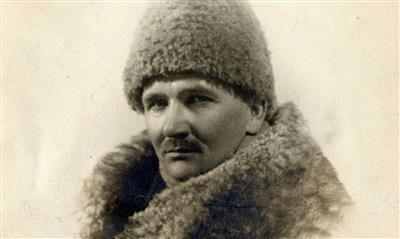}
        \centerline{(b) Artist: Sallinen Tyko}
    \end{minipage}
    \caption{The artwork Ruokokoski Maalaa (a) presented alongside an image of its creator, Sallinen Tyko (b). This illustrates the real-world visual and historical context underlying the dataset.}
    \label{fig:wikiart_example}
\end{figure}

A visual example of a painting and its corresponding author is illustrated in Figure \ref{fig:wikiart_example}.

\subsection{Ontological Schema and Neo4j Modeling}
To move beyond flat, tabular metadata, all extracted entities and relationships were instantiated in Neo4j, a native graph database optimised for traversing highly interconnected data. Modeling the domain as a network of nodes (entities) connected by directed edges (relationships) allows us to explicitly capture the historical lineage of the training data. 

\begin{figure}[htbp]
    \centering
    \includegraphics[width=0.7\linewidth]{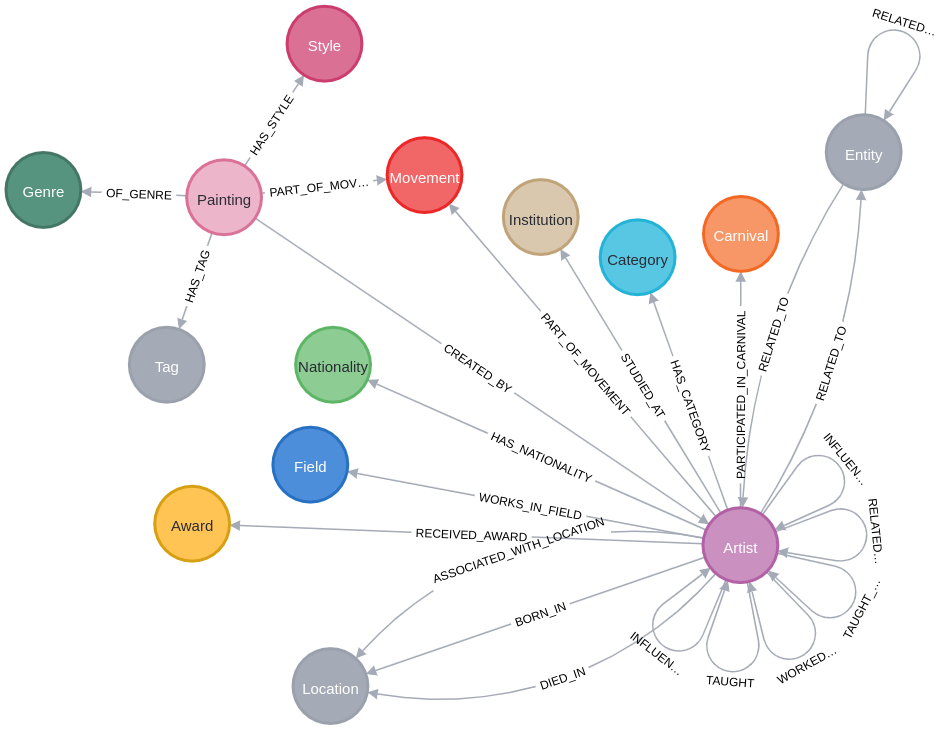}
    \caption{Ontological schema of the Artistic KG ($\mathcal{G}_D$), defining the structural relationships used to map ArtBench to its domain context.}
    \label{fig:kg_schema}
\end{figure}

The ontological schema for our artistic domain centers on two primary anchor nodes: \texttt{Painting} and \texttt{Artist}. Around these central nodes, we linked various contextual entities. The overall structural design of this ontology is illustrated in Figure \ref{fig:kg_schema}.

By mapping the extracted data to this schema, we ensure that the resulting topology accurately reflects ground-truth art history. Structuring the domain in Neo4j provides the necessary foundation for our problem. For example, if two visually distinct paintings were created by artists who studied under the same master, the graph structure connects them via a short path length, allowing our embedding algorithms to capture their structural proximity.

\paragraph{Relational Logic.} Table \ref{tab:kg_relationships} outlines the core deterministic relationships utilised to construct the graph. This schema defines how the framework traverses the space between a physical artifact and its historical context.

\begin{table}[ht]
\centering
\caption{Core Entity Relationships within the Artistic Knowledge Graph ($\mathcal{G}_D$). This schema defines the permissible edges that control the structural topology of the domain.}
\label{tab:kg_relationships}
\footnotesize
\begin{tabular}{@{}lll@{}}
\toprule
\textbf{Source Node} & \textbf{Relationship Edge} & \textbf{Target Node} \\ \midrule
\texttt{Painting}    & \texttt{CREATED\_BY}       & \texttt{Artist}      \\
\texttt{Painting}    & \texttt{BELONGS\_TO\_STYLE}& \texttt{ArtStyle}    \\
\texttt{Painting}    & \texttt{HAS\_GENRE}        & \texttt{Genre}       \\
\texttt{Painting}    & \texttt{USES\_MEDIUM}      & \texttt{Medium}      \\
\texttt{Artist}      & \texttt{INFLUENCED\_BY}    & \texttt{Artist}      \\
\texttt{Artist}      & \texttt{STUDIED\_UNDER}    & \texttt{Artist}      \\
\texttt{Artist}      & \texttt{ASSOCIATED\_WITH}  & \texttt{ArtMovement} \\
\texttt{Artist}      & \texttt{BORN\_IN}          & \texttt{Location}    \\
\bottomrule
\end{tabular}
\end{table}

\paragraph{Scale and Connectivity.} To evaluate the complexity and semantic density of our database, we provide a quantitative inventory in Table \ref{tab:art_kg_inventory}. The high volume of unique tags and artist relationships ensures that the graph defines a rich semantic space, preventing the attribution from collapsing into generic clusters.

\begin{table}[ht]
\centering
\caption{KG Inventory: Quantitative Analysis of Relationship Density and Semantic Diversity across $\mathcal{G}_D$.}
\label{tab:art_kg_inventory}
\small
\begin{tabular}{llrrr}
\toprule
\textbf{Source} & \textbf{Relationship} & \textbf{Target} & \textbf{Count} & \textbf{Unique} \\
\midrule
Painting & HAS\_TAG & Tag & 122448 & 3651 \\
Painting & CREATED\_BY & Artist & 47224 & 1769 \\
Painting & HAS\_STYLE & Style & 47224 & 81 \\
Painting & OF\_GENRE & Genre & 47224 & 339 \\
Painting & PART\_OF\_MOVEMENT & Movement & 47224 & 73 \\
Artist & HAS\_CATEGORY & Category & 37483 & 9776 \\
Artist & ASSOCIATED\_WITH\_LOCATION & Location & 4147 & 2284 \\
Artist & WORKS\_IN\_FIELD & Field & 3597 & 241 \\
Artist & PART\_OF\_MOVEMENT & Movement & 3575 & 393 \\
Artist & STUDIED\_AT & Institution & 3064 & 701 \\
Entity & RELATED\_TO & Entity & 2554 & 960 \\
Artist & HAS\_NATIONALITY & Nationality & 2113 & 93 \\
Artist & BORN\_IN & Location & 1695 & 1018 \\
Artist & DIED\_IN & Location & 1456 & 627 \\
Artist & INFLUENCED\_BY & Artist & 796 & 321 \\
Artist & WORKED\_WITH & Artist & 652 & 350 \\
Artist & INFLUENCED & Artist & 627 & 362 \\
Artist & RECEIVED\_AWARD & Award & 295 & 231 \\
Artist & TAUGHT\_BY & Artist & 291 & 171 \\
Artist & TAUGHT & Artist & 250 & 214 \\
Artist & RELATED\_TO & Artist & 121 & 104 \\
Artist & PARTICIPATED\_IN\_CARNIVAL & Carnival & 2 & 2 \\
\bottomrule
\end{tabular}
\end{table}

\paragraph{Metadata Details.} The internal properties of each node type are detailed in Table \ref{tab:art_node_edge_properties}. These fields allow for detailed filtering, such as isolating artists active during specific decades, which is critical for the mathematical sharpening of the attribution scores.

\begin{figure}[htbp]
    \centering
    \begin{minipage}{0.45\linewidth}
        \centering
        \includegraphics[width=\linewidth]{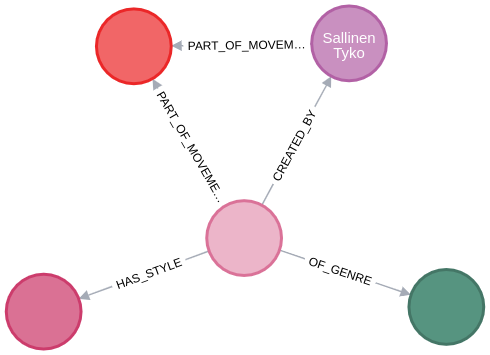}
        \centerline{(a) Artwork: Ruokokoski Maalaa}
    \end{minipage}\hfill
    \begin{minipage}{0.50\linewidth}
        \centering
        \includegraphics[width=\linewidth]{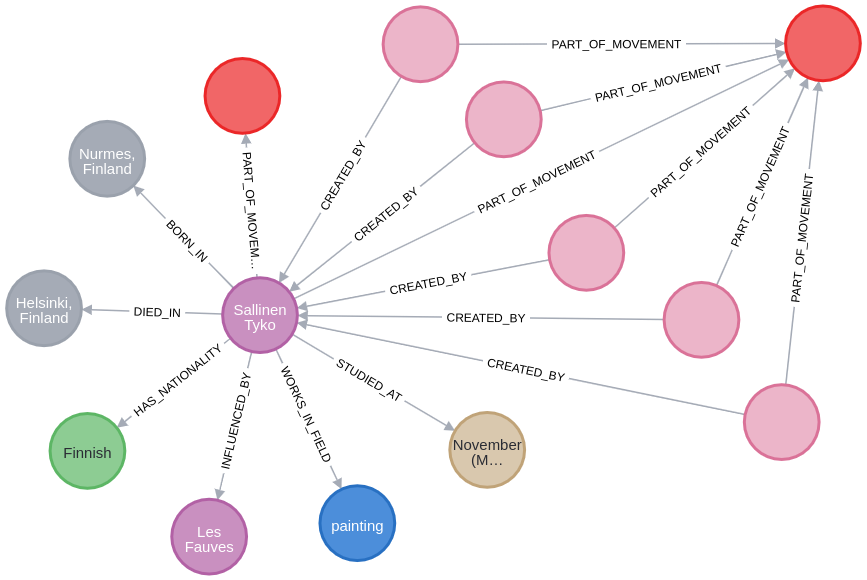}
        \centerline{(b) Artist: Sallinen Tyko}
    \end{minipage}
    \caption{Graph visualisation of the painting Ruokokoski Maalaa (a) and the artist Sallinen Tyko (b). The network illustrates the central Artist and Painting nodes, detailing their surrounding connections by displaying a subset of the artist's portfolio alongside broader metadata.}
    \label{fig:wikiart_kg_example}
\end{figure}

In Figure \ref{fig:wikiart_kg_example} we can see the KG representation of a painting and its corresponding author (the same as in Figure \ref{fig:wikiart_example}) in Neo4j Database.

\begin{table}[ht]
\centering
\caption{Internal Node Property Schema: Detailed metadata fields stored within the $\mathcal{G}_D$ topology for filtering and state representation.}
\label{tab:art_node_edge_properties}
\small
\begin{tabular}{ll}
\toprule
\textbf{Node Label} & \textbf{Internal Properties (Metadata Fields)} \\
\midrule
Artist & id, name, born\_date, died\_date, wikipedia\_summary, active\_years \\
Genre & id, display\_name \\
MainTarget & id, url \\
Movement & id, display\_name \\
Painting & id, url \\
Style & id, display\_name \\
Tag & id, display\_name \\
\bottomrule
\end{tabular}
\end{table}

\section{Contrastive Learning and Representation Alignment}
\label{app:contrastive_learning}

To compute the Bayesian attribution components in the Domain Space, the feature Likelihood $P(\mathbf{C} \mid \mathbf{X}_j)$ and the Prior $P(\mathbf{X}_j)$, the framework requires a unified semantic space. As discussed in Section \ref{sec:art_datasets}, generative models synthesise novel artifacts ($\mathbf{C}$) that lack intrinsic metadata. To evaluate these outputs against our Domain KG, we establish a mathematical bridge between continuous visual features and discrete structural knowledge. 

Our approach utilises a cross-modal contrastive learning objective. Inspired by the CLIP architecture \cite{clip_contrastive}, we align the latent distributions of two different modalities into a shared vector space, allowing for direct cosine similarity measurements between pixels and graph nodes.

\subsection{Encoder Architectures and Domain Extraction}
To construct this joint space, we must first extract unimodal representations from both the visual and structural domains. We carefully selected state-of-the-art encoders optimised for their respective modalities.

\paragraph{Visual Encoding.} 
For the continuous image modality, we utilise the ViT-g-14 architecture \cite{openclip}. We selected this model because its large-scale pre-training provides the discriminative depth necessary to capture stylistic characteristics that smaller encoders might overlook. This choice is further validated in our ablation study (see Table \ref{tab:encoder_ablation_combined}), where the ViT-g-14 demonstrated superior predictive faithfulness compared to alternative architectures. In the context of attribution, this high-dimensional capacity is essential for distinguishing between closely related training samples that share surface-level similarities but differ in stylistic characteristics.

\paragraph{Structural Encoding via Neo4j and Node2Vec.} 
The domain metadata is stored natively in Neo4j and encoded using the Node2Vec algorithm \cite{node2vec}. Node2Vec’s biased random walks allow us to tune the representation between local stylistic details and global historical context. A key advantage of this approach is its seamless integration with the Neo4j Graph Data Science (GDS) library, which provides optimised, production-ready procedures to extract high-dimensional embeddings directly within the database environment. This efficient implementation ensures that the structural prior $P(\mathbf{X}_j)$ accurately reflects the topological rarity of a node’s position within the art-historical network, preventing generic hubs, such as heavily represented styles, from dominating the attribution results.

\subsection{Alignment Procedure and InfoNCE Loss}
\label{app:contrastive_procedure}

To construct the joint multimodal manifold, raw representations from the ViT and Node2Vec encoders are passed through learned non-linear projection heads. This projection is necessary to map disparate latent distributions into a shared $d_e$-dimensional space ($d_e = 512$). The procedure is governed by a Symmetric InfoNCE Loss, mirroring the contrastive objective utilised in the CLIP architecture \cite{clip_contrastive}, that maximises the lower bound of the mutual information between visual textures and structural graph signatures. By adopting this loss, we ensure that the model learns a discriminative representation where matched pairs are strongly attracted while mismatched pairs are effectively repelled in the joint embedding space.

\paragraph{Architectural Specifics.} 
Each projection head consists of a deep Multi-Layer Perceptron (MLP) designed for high-fidelity alignment. To prevent mode collapse, we incorporate Batch Normalization layers after the first linear transformation. Additionally, a dropout rate of $0.2$ is applied to enhance generalization and prevent the model from over-fitting to specific historical samples in the training set. A critical component of our architecture is the Dual-Normalization strategy: input vectors are L2-normalised before the MLP to stabilise the initial gradients, and final embeddings are re-normalised to ensure they lie on a hypersphere, making the cosine similarity calculation mathematically consistent.

\paragraph{Optimization and Training Schedule.} 
The alignment model is optimised using the AdamW optimiser with a weight decay of $10^{-4}$ to ensure weight sparsity. We utilise a Cosine Annealing learning rate scheduler, which starts at a conservative $10^{-4}$ and smoothly decays to zero over $200$ epochs. This schedule allows the model to explore the complex cross-modal landscape early in training while finely tuning the alignment in the final stages. To calibrate the similarity scores, we employ a learnable temperature parameter $\tau$, initialised at $0.07$ and clamped at $100$ to prevent gradient explosions during the calculation of logits.

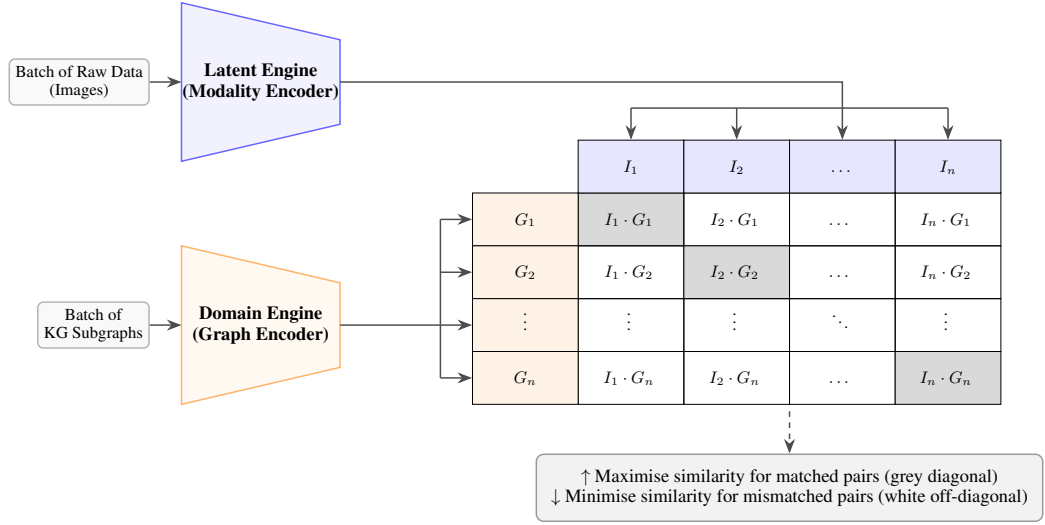
\begin{figure}[htbp]
\centering
\resizebox{0.98\textwidth}{!}{%
\begin{tikzpicture}[
    box/.style={draw, rectangle, rounded corners=3pt, align=center, thick, font=\small, fill=gray!5, draw=gray!60},
    arrow/.style={-{Stealth[scale=1.2]}, thick, draw=black!70},
    cell/.style={draw, rectangle, minimum width=2.0cm, minimum height=1.0cm, align=center, font=\small, anchor=center, text height=1.5ex, text depth=0.25ex},,
    poscell/.style={cell, fill=gray!30, font=\small\bfseries}, 
    embed_top/.style={cell, fill=blue!10},
    embed_left/.style={cell, fill=orange!10}
]

\matrix (sim_matrix) [matrix of nodes, ampersand replacement=\&, column sep=-\pgflinewidth, row sep=-\pgflinewidth] {
    |[poscell]| $I_1 \cdot G_1$ \& |[cell]| $I_2 \cdot G_1$ \& |[cell]| $\dots$ \& |[cell]| $I_n \cdot G_1$ \\
    |[cell]| $I_1 \cdot G_2$ \& |[poscell]| $I_2 \cdot G_2$ \& |[cell]| $\dots$ \& |[cell]| $I_n \cdot G_2$ \\
    |[cell]| $\vdots$ \& |[cell]| $\vdots$ \& |[cell]| $\ddots$ \& |[cell]| $\vdots$ \\
    |[cell]| $I_1 \cdot G_n$ \& |[cell]| $I_2 \cdot G_n$ \& |[cell]| $\dots$ \& |[poscell]| $I_n \cdot G_n$ \\
};

\node (i1) [embed_top, above=-\pgflinewidth of sim_matrix-1-1] {$I_1$};
\node (i2) [embed_top, above=-\pgflinewidth of sim_matrix-1-2] {$I_2$};
\node (idots) [embed_top, above=-\pgflinewidth of sim_matrix-1-3] {$\dots$};
\node (in) [embed_top, above=-\pgflinewidth of sim_matrix-1-4] {$I_n$};

\node (g1) [embed_left, left=-\pgflinewidth of sim_matrix-1-1] {$G_1$};
\node (g2) [embed_left, left=-\pgflinewidth of sim_matrix-2-1] {$G_2$};
\node (gdots) [embed_left, left=-\pgflinewidth of sim_matrix-3-1] {$\vdots$};
\node (gn) [embed_left, left=-\pgflinewidth of sim_matrix-4-1] {$G_n$};


\coordinate (E2_out) at ([xshift=-2.5cm]gdots.west);
\coordinate (E2_in)  at ([xshift=-3.0cm]E2_out);

\coordinate (E2_TL) at ([yshift=1.5cm]E2_in);
\coordinate (E2_BL) at ([yshift=-1.5cm]E2_in);
\coordinate (E2_TR) at ([yshift=0.8cm]E2_out);
\coordinate (E2_BR) at ([yshift=-0.8cm]E2_out);

\draw [fill=orange!5, draw=orange!60, thick] (E2_TL) -- (E2_TR) -- (E2_BR) -- (E2_BL) -- cycle;
\node at ([xshift=-1.5cm]E2_out) [align=center, font=\bfseries] {Domain Engine\\(Graph Encoder)};

\coordinate (E1_out) at ([yshift=4.6cm]E2_out);
\coordinate (E1_in)  at ([xshift=-3.0cm]E1_out);

\coordinate (E1_TL) at ([yshift=1.5cm]E1_in);
\coordinate (E1_BL) at ([yshift=-1.5cm]E1_in);
\coordinate (E1_TR) at ([yshift=0.8cm]E1_out);
\coordinate (E1_BR) at ([yshift=-0.8cm]E1_out);

\draw [fill=blue!5, draw=blue!60, thick] (E1_TL) -- (E1_TR) -- (E1_BR) -- (E1_BL) -- cycle;
\node at ([xshift=-1.5cm]E1_out) [align=center, font=\bfseries] {Latent Engine\\(Modality Encoder)};

\node (graph_in) [box, left=0.6cm of E2_in] {Batch of\\KG Subgraphs};
\node (latent_in) [box, left=0.6cm of E1_in] {Batch of Raw Data\\(Images)};

\draw [arrow] (graph_in.east) -- (E2_in);
\draw [arrow] (latent_in.east) -- (E1_in);


\coordinate (branch_h_center) at ([yshift=0.6cm]idots.north);
\draw [thick, draw=black!70] (E1_out) -| (branch_h_center); 

\draw [thick, draw=black!70] ([yshift=0.6cm]i1.north) -- ([yshift=0.6cm]in.north);
\draw [arrow] ([yshift=0.6cm]i1.north) -- (i1.north);
\draw [arrow] ([yshift=0.6cm]i2.north) -- (i2.north);
\draw [arrow] ([yshift=0.6cm]idots.north) -- (idots.north);
\draw [arrow] ([yshift=0.6cm]in.north) -- (in.north);

\coordinate (branch_v_center) at ([xshift=-0.6cm]gdots.west);
\draw [thick, draw=black!70] (E2_out) -- (branch_v_center);

\draw [thick, draw=black!70] ([xshift=-0.6cm]g1.west) -- ([xshift=-0.6cm]gn.west);
\draw [arrow] ([xshift=-0.6cm]g1.west) -- (g1.west);
\draw [arrow] ([xshift=-0.6cm]g2.west) -- (g2.west);
\draw [arrow] ([xshift=-0.6cm]gdots.west) -- (gdots.west);
\draw [arrow] ([xshift=-0.6cm]gn.west) -- (gn.west);

\node (loss_box) [below=0.8cm of sim_matrix, draw=gray!80, fill=gray!10, thick, rounded corners=4pt, align=center, inner sep=8pt] {
    $\uparrow$ Maximise similarity for matched pairs (grey diagonal) \\
    $\downarrow$ Minimise similarity for mismatched pairs (white off-diagonal)
};

\draw [arrow, dashed] (sim_matrix.south) -- (loss_box.north);

\end{tikzpicture}%
} 
\caption{The contrastive multimodal alignment procedure. A batch of raw continuous data (images) and their corresponding subgraphs are passed through parallel stacked encoders. A symmetric InfoNCE loss is applied to the resulting similarity matrix to maximise the alignment of corresponding pairs (the highlighted diagonal) while pushing apart mismatched pairs.}
\label{fig:contrastive_training}
\end{figure}

The contrastive training procedure illustrated in Figure \ref{fig:contrastive_training} establishes the shared representation space required to perform direct semantic comparisons between generated artifacts and the training dataset.

The alignment objective is explicitly optimised through a batch-wise similarity calculation scaled by a learnable temperature parameter $\tau$. This mechanism ensures that the model remains discriminative across a diverse range of artistic styles and historical contexts. The complete algorithmic implementation of this alignment loop, utilising the CLIP-style symmetric loss, is detailed in Figure \ref{fig:contrastive_pseudocode}.

\begin{figure}[htpb]
\begin{lstlisting}[language=Python, basicstyle=\ttfamily\footnotesize, frame=single]
# image_encoder - ViT-g-14
# graph_encoder - Node2Vec (extracted via Neo4j)
# I[n, c, h, w] - minibatch of raw training images
# G[n, d_g]     - minibatch of topological graph embeddings
# MLP_i         - non-linear projection head for visual features
# MLP_g         - non-linear projection head for graph features
# t             - learned temperature parameter

# 1. Extract feature representations of each modality
I_f = image_encoder(I) #[n, d_i]
G_f = graph_encoder(G) #[n, d_g]

# 2. Project into joint multimodal embedding space [n, d_e]
# Output vectors are strictly L2-normalized within the MLP heads
I_e = MLP_i(I_f)
G_e = MLP_g(G_f)

# 3. Compute scaled pairwise cosine similarities [n, n]
logits = np.dot(I_e, G_e.T) * np.exp(t)

# 4. Compute symmetric loss function (CLIP Objective)
labels = np.arange(n)
loss_i = cross_entropy_loss(logits, labels, axis=0)
loss_g = cross_entropy_loss(logits, labels, axis=1)
loss = (loss_i + loss_g) / 2
\end{lstlisting}
\caption{Algorithmic implementation of the contrastive alignment loop. The procedure leverages parallel projection heads and a symmetric loss function to maximise the cross-modal mutual information required for grounded Bayesian attribution.}
\label{fig:contrastive_pseudocode}
\end{figure}

\section{Generative Setup and Hyperparameters}
\label{app:hyperparameters}

To ensure an exact and rigorous comparison with our parametric baselines, we strictly replicate the diffusion training and generation pipeline utilised by D-TRAK \cite{d-trak} for the ArtBench dataset \cite{artbench}. 

\paragraph{Model Architecture and Fine-Tuning.}
We utilise a pre-trained Stable Diffusion model adapted to a $256 \times 256$ resolution. To make training computationally tractable while preserving generative quality, we employ Low-Rank Adaptation (LoRA) on the cross-attention layers of the U-Net. The LoRA rank dimension is set to $r=128$, resulting in approximately 25.5M trainable parameters. The model is fine-tuned conditionally using textual prompts formatted as ``a \{class\} painting'' (e.g., ``a ukiyo-e painting''). 

\paragraph{Optimization Schedule.}
The model is trained for 100 epochs using a total batch size of 64. We optimise the parameters using the AdamW optimiser with a weight decay of $10^{-6}$ and a dropout rate of 0.1. Data augmentation is limited to random horizontal flips. The learning rate follows a cosine annealing schedule, initiating with a 0.1 fraction warmup to a peak learning rate of $3 \times 10^{-4}$. 

\paragraph{Inference and Generation.}
During the evaluation phase, target images ($\mathbf{C}$) are synthesised utilising the 50-step Denoising Diffusion Implicit Model (DDIM) solver. To enforce strong class conditioning, we apply classifier-free guidance with a fixed scale of 7.5.

\section{Linear Datamodeling Score (LDS) Formulation}
\label{app:lds_math}

\subsection{Conceptual Framework}
The Linear Datamodeling Score (LDS), introduced by \cite{park2023trak}, formalises the evaluation of data attribution as a counterfactual prediction task. The core objective is to determine if an attribution method can accurately predict how a model's output would change if specific training samples were removed. 

For a fixed target $z$, the model output $f(z; \theta^*(S'))$, where $\theta^*(S')$ represents the parameters resulting from training on a subset $S' \subset S$, is viewed as a function of the training subset. An ideal attribution method $\tau$ should provide scores that, when summed, approximate the model's behavior on any given subset.

\subsection{Mathematical Definition}
Following the formulation of the original TRAK framework \cite{park2023trak}, the attribution-based prediction for a subset $S_j$ is defined as the additive sum of the individual scores of all training samples $z_i$ contained within that subset:
\begin{equation}
    g_\tau(z, S_j) = \sum_{i: z_i \in S_j} \tau(z, S)_i
\end{equation}

The LDS for a specific target $z$ is then defined as the Spearman rank correlation ($\rho$) between the actual model outputs across $m$ randomly sampled subsets and the corresponding attribution-based predictions:
\begin{equation}
    \text{LDS}(\tau, z) := \rho \left( \{ f(z; \theta^*(S_j)) \}_{j=1}^m, \{ g_\tau(z, S_j) \}_{j=1}^m \right)
\end{equation}
where $\rho$ denotes the Spearman rank correlation. A high correlation indicates that the attribution scores $\tau$ correctly capture the marginal contribution of each training sample to the final output.

\subsection{Implementation for Diffusion Models}
Following the rigorous protocol established by D-TRAK \cite{d-trak}, we implement several approximations to account for the stochasticity inherent in generative diffusion processes. We first perform \textbf{ensemble training} by sampling $M = 64$ random subsets of the training set $\mathcal{D}$, each consisting of a fraction $\alpha = 0.5$ of the total data. To mitigate variance from random initialization, we train an ensemble of three independent models with different random seeds for every subset. To evaluate the resulting model output $f(z; \theta^*(S_j))$, we compute the diffusion loss $L_{\text{Simple}}$ using a robust \textbf{loss approximation} strategy. Specifically, to ensure a stable estimate of the expectation over time ($\mathbb{E}_t$) and noise ($\mathbb{E}_\epsilon$), we evaluate the loss across 1,000 evenly spaced timesteps within the interval $[1, T]$, sampling three standard Gaussian noise vectors $\epsilon \sim \mathcal{N}(0, \mathbf{I})$ at each step. The \textbf{final scoring} is determined by calculating the LDS for each sample of interest across both the validation and generated sets, with the final reported metric being the arithmetic mean of these individual Spearman correlations.

\subsection{Statistical Significance and Uncertainty Quantification}
\label{app:stat_sign}
To ensure the reported LDS metrics are statistically robust and not biased by specific subset selections, we perform uncertainty quantification following the methodology of prior work. We calculate 95\% confidence intervals for all reported correlations using a bootstrapping procedure with $n=64$ resamples. For each resample, we compute the mean Spearman rank correlation across our evaluation sets. This process allows us to report both the stability and the significance of our attribution results, ensuring that the performance gains of our framework are consistent across different random draws of the training distribution.

\section{Baseline Implementations}
\label{app:baselines}

We focus on conducting data attribution in a post-hoc manner, which refers to
the application of attribution methods after model training. Post-hoc attribution methods do not add extra restrictions to how we train models and are thus preferred in practice \cite{ribeiro2016model}.

\paragraph{\textbf{Textual Similarity.}} This baseline measures the semantic overlap between the generation prompt and the training captions. In the ArtBench dataset, training captions follow a standardised template: \textit{"a \{style\} painting."} Similarly, our generation prompts utilised the same class-conditional format. To compute this score, we utilise a Jaccard Similarity metric on the tokenised keywords of the strings. 

\paragraph{\textbf{Raw Pixel.}} As a naive image-space baseline, this method computes the direct mathematical similarity between the raw pixel values of the images. We represent both the generated artifact $\mathbf{C}$ and the training samples $\mathbf{X}_j$ as flattened RGB vectors. We then calculate the dot product and cosine similarity between these vectors to measure spatial and coloristic overlap. 

\paragraph{\textbf{CLIP Similarity.}} This baseline serves as a standard similarity-based attribution method by encoding both the generated target and the training samples into a shared multimodal space using a pre-trained CLIP vision encoder \cite{clip_contrastive}. Following established protocols for black-box data attribution, the influence score is calculated as the dot product or cosine similarity between these dense embeddings. While CLIP remains the primary similarity-based reference, we also investigated the impact of various alternative vision encoders on attribution performance. A comprehensive comparison of these architectures and their respective results is detailed in our ablation study, specifically provided in Table \ref{tab:encoder_ablation_combined}.

\paragraph{\textbf{Gradient.}}  This estimator from Charpiat et al. \cite{charpiat2019input} computes the direct dot product or cosine similarity using the gradient representation of the generated sample and the gradient representation of the training sample.

\paragraph{\textbf{Gradient Similarity.}}  Following the approach of Charpiat et al. \cite{charpiat2019input}, this gradient-based estimator quantifies influence by measuring the alignment between projected gradient representations. It captures input similarity from the perspective of the model's internal parameters using the following dot product or cosine similarity formulations:
\begin{equation}
\tau(x, \mathcal{D})_n = \mathcal{P}^\top \nabla_\theta \mathcal{L}_{\text{Simple}}(x; \theta^*)^\top \cdot \mathcal{P}^\top \nabla_\theta \mathcal{L}_{\text{Simple}}(x^n; \theta^*)
\end{equation}
\begin{equation}
\tau(x, \mathcal{D})_n = \frac{\mathcal{P}^\top \nabla_\theta \mathcal{L}_{\text{Simple}}(x; \theta^*)^\top \cdot \mathcal{P}^\top \nabla_\theta \mathcal{L}_{\text{Simple}}(x^n; \theta^*)}{||\mathcal{P}^\top \nabla_\theta \mathcal{L}_{\text{Simple}}(x; \theta^*)|| \cdot ||\mathcal{P}^\top \nabla_\theta \mathcal{L}_{\text{Simple}}(x^n; \theta^*)||}
\end{equation}

\paragraph{\textbf{TracInCP.}} We utilise the TracInCP estimator \cite{pruthi2020}, which identifies training data influence by tracing the model's performance change along the learning trajectory. Influence is approximated by averaging the inner products of the projected gradients at $C$ evenly spaced checkpoints $\theta^c$:
\begin{equation}
\tau(x, \mathcal{D})_n = \frac{1}{C} \sum_{c=1}^{C} \mathcal{P}^\top_c \nabla_\theta \mathcal{L}_{\text{Simple}}(x; \theta^c)^\top \cdot \mathcal{P}^\top_c \nabla_\theta \mathcal{L}_{\text{Simple}}(x^n; \theta^c)
\end{equation}

\paragraph{\textbf{GAS.}} The Gradient Averaging Similarity (GAS) is a variant of the TracInCP method that normalises the gradient vectors to calculate cosine similarity rather than raw dot products \cite{hammoudeh2022identifying}.

\paragraph{\textbf{TRAK.}} 
We adapt the standard TRAK estimator \cite{park2023trak} to the diffusion setting. It applies random projections to the model's gradients to create a tractable, low-dimensional gradient matrix, calculating attribution via a regularised kernel dot product.

\begin{itemize}
    \item Retraining-free TRAK, where a damping term $\lambda I$ is included to ensure numerical stability and provide regularization:
    \begin{equation}
        \Phi_{\text{TRAK}} = [\phi(x^1); \dots; \phi(x^N)]^\top
    \end{equation}
    \begin{equation}
        \tau(x, \mathcal{D})_n = \mathcal{P}^\top \nabla_\theta \mathcal{L}_{\text{Simple}}(x; \theta^*)^\top \cdot (\Phi_{\text{TRAK}}^\top \Phi_{\text{TRAK}} + \lambda I)^{-1} \cdot \mathcal{P}^\top \nabla_\theta \mathcal{L}_{\text{Simple}}(x^n; \theta^*)
    \end{equation}

    \item Ensemble TRAK: To improve robustness, the ensemble variant averages influence scores across $S$ models trained on distinct data subsets:
    \begin{equation}
        \Phi_{\text{TRAK}}^s = [\phi^s(x^1); \dots; \phi^s(x^N)]^\top, \quad \text{where } \phi^s(x) = \mathcal{P}_s^\top \nabla_\theta \mathcal{L}_{\text{Simple}}(x; \theta_s^*)
    \end{equation}
    \begin{equation}
        \tau(x, \mathcal{D})_n = \frac{1}{S} \sum_{s=1}^S \mathcal{P}_s^\top \nabla_\theta \mathcal{L}_{\text{Simple}}(x; \theta_s^*)^\top \cdot (\Phi_{\text{TRAK}}^{s \top} \Phi_{\text{TRAK}}^s + \lambda I)^{-1} \cdot \mathcal{P}_s^\top \nabla_\theta \mathcal{L}_{\text{Simple}}(x^n; \theta_s^*)
    \end{equation}
\end{itemize}

\paragraph{\textbf{Relative Influence.}} The Relative Influence, introduced by Barshan et al. \cite{barshan2020relatif}, seeks to improve the standard influence function estimator by normalising it against the magnitude of the Hessian-Vector Product (HVP). We adapt this methodology to the diffusion setting and apply the same scalability optimizations used in TRAK. The attribution score is calculated as:

\begin{equation}
\tau(x, \mathcal{D})_n = \frac{\mathcal{P}^\top \nabla_\theta \mathcal{L}_{\text{Simple}}(x; \theta^*)^\top \cdot (\Phi_{\text{TRAK}}^\top \Phi_{\text{TRAK}} + \lambda I)^{-1} \cdot \mathcal{P}^\top \nabla_\theta \mathcal{L}_{\text{Simple}}(x^n; \theta^*)}{||(\Phi_{\text{TRAK}}^\top \Phi_{\text{TRAK}} + \lambda I)^{-1} \cdot \mathcal{P}^\top \nabla_\theta \mathcal{L}_{\text{Simple}}(x^n; \theta^*)||}
\end{equation}

\paragraph{Renormalised Influence.}
Introduced by Hammoudeh and Lowd \cite{hammoudeh2022identifying}, the Renormalised Influence baseline seeks to calibrate influence scores by normalising them against the magnitude of the training sample's gradients. The resulting renormalised attribution score is formulated as:
\begin{equation}
\tau(x, \mathcal{D})_n = \frac{\mathcal{P}^\top \nabla_\theta \mathcal{L}_{\text{Simple}}(x; \theta^*)^\top \cdot (\Phi_{\text{TRAK}}^\top \Phi_{\text{TRAK}} + \lambda I)^{-1} \cdot \mathcal{P}^\top \nabla_\theta \mathcal{L}_{\text{Simple}}(x^n; \theta^*)}{||\mathcal{P}^\top \nabla_\theta \mathcal{L}_{\text{Simple}}(x^n; \theta^*)||}
\end{equation}

\paragraph{\textbf{Journey TRAK.}} 
Georgiev et al. \cite{georgiev2023journey} introduced Journey TRAK, which originally focuses on attributing the intermediate noisy images $x_t$ along the diffusion trajectory rather than the final generated output. The formulation is as follows:

\begin{equation}
\tau(x, \mathcal{D})_n = \frac{1}{T'} \sum_{t=1}^{T'} \mathcal{P}^\top \nabla_\theta \mathcal{L}_{\text{Simple}}^t(x_t; \theta^*)^\top \cdot (\Phi_{\text{TRAK}}^\top \Phi_{\text{TRAK}} + \lambda I)^{-1} \cdot \mathcal{P}^\top \nabla_\theta \mathcal{L}_{\text{Simple}}(x^n; \theta^*)
\end{equation}

where $T'$ represents the total number of inference steps (set to 50 here), and $x_t$ denotes the noisy generated image at timestep $t$ along the sampling trajectory.

\paragraph{\textbf{D-TRAK.}} 
Similar to TRAK, we adapt D-TRAK\cite{d-trak} to our setting. We implement the model output function $f(z, \theta)$ as $\mathcal{L}_{\text{Square}}$. D-TRAK is implemented using the following equations:

\begin{equation}
\boldsymbol{\Phi}_{\text{D-TRAK}} = [\boldsymbol{\Phi}(\boldsymbol{x}^1), \dots, \boldsymbol{\Phi}(\boldsymbol{x}^N)]^\top, \quad \text{where } \boldsymbol{\Phi}(\boldsymbol{x}) = \boldsymbol{P}^\top \nabla_\theta \mathcal{L}_{\text{Simple}}(\boldsymbol{x}, \theta)
\end{equation}

\begin{equation}
\tau(z, \mathbb{S})^i = (\boldsymbol{P}^\top \nabla_\theta \mathcal{L}_{\text{Simple}}(\boldsymbol{x}, \theta))^\top \cdot \left(\boldsymbol{\Phi}_{\text{TRAK}}^\top \boldsymbol{\Phi}_{\text{TRAK}} + \lambda \boldsymbol{I}\right)^{-1} \cdot \boldsymbol{P}^\top \nabla_\theta \mathcal{L}_{\text{Simple}}(\boldsymbol{x}^i, \theta)
\end{equation}

where $\lambda \boldsymbol{I}$ is also included for numerical stability and regularization, as in TRAK. Additionally, the output function $f(z, \theta)$ could be replaced with other functions.

\paragraph{\textbf{Diffusion Attribution Score (DAS).}}
Introduced by Lin et al. \cite{das}, DAS addresses the limitations of standard loss-based attribution. It directly compares the predicted distributions using Kullback-Leibler (KL) divergence. To isolate the effect of removing a specific training sample $z^{(i)}$ on the generated target $z_{\text{gen}}$, the score is defined mathematically as the KL-divergence between the original and retrained models' noise predictors:
\begin{equation}
    \tau_{\text{DAS}}(z_{\text{gen}}, \mathcal{S})^{(i)} \approx \mathbb{E}_{\epsilon,t} \left[ ||\epsilon_\theta(x_t^{\text{gen}}, t) - \epsilon_{\theta \backslash i}(x_t^{\text{gen}}, t)||^2 \right]
\end{equation}

DAS estimates the change in model parameters using a single Newton step and also normalises and averages the gradients and residuals over the entire generation trajectory, and utilises random projection matrices ($\mathcal{P}$) to compress the gradient dimension. 

Let $g(x)$ denote the time-averaged, normalised, and randomly projected gradient of the noise predictor, $G(\mathcal{S})$ represent the stacked gradient matrix for the training set, and $r^{(i)}$ denote the averaged residual. The final, tractable DAS formulation is computed as:
\begin{equation}
    \tau_{\text{DAS}}(z_{\text{gen}}, \mathcal{S})^{(i)} = \left\| \frac{ g(x_{\text{gen}})^\top \left[ G(\mathcal{S})^\top G(\mathcal{S}) \right]^{-1} g(x^{(i)}) r^{(i)} }{ 1 - g(x^{(i)})^\top \left[ G(\mathcal{S})^\top G(\mathcal{S}) \right]^{-1} g(x^{(i)}) } \right\|^2
\end{equation}

\section{Ablation Study}

In this section, we provide a detailed analysis of the architectural and mathematical choices that define our framework. We first examine how the choice of visual encoder affects attribution accuracy. We then evaluate the impact of our Bayesian components, specifically the non-linear scaling, the dataset prior, and our fusion strategy, to show how each part contributes to the final performance.

\paragraph{Visual Encoder Scaling and Data Diversity}: As shown in Table \ref{tab:encoder_ablation_combined}, we tested several vision backbones to find the best feature extractor for our Latent Space. We found a clear link between the model size and the accuracy of the attribution results. While smaller models like ViT-B/32 struggle to find the specific styles needed for ArtBench, the ViT-g-14 model from OpenCLIP consistently gives the highest results. This performance is likely because ViT-g-14 was trained on the massive LAION-2B dataset. This large-scale, high-diversity training set allows the model to build a more general understanding of images, which is necessary for tracking influences across complex domains.

\begin{table*}[ht]
\centering
\caption{Ablation study of visual encoders. We report the LDS, \% for all the selected encoders. The ViT-g-14 model consistently provides the best results across both datasets. This is likely due to the massive scale and diversity of the LAION-2B dataset used for its pre-training, which allows the model to understand complex artistic styles better than smaller-scale alternatives.}
\label{tab:encoder_ablation_combined}
\footnotesize
\resizebox{\textwidth}{!}{%
\begin{tabular}{lcccc}
\toprule
 & \multicolumn{2}{c}{\textbf{ArtBench-2 LDS (\%)}} & \multicolumn{2}{c}{\textbf{ArtBench-5 LDS (\%)}} \\
\cmidrule(lr){2-3} \cmidrule(lr){4-5}
\textbf{Encoder Architecture} & \textbf{Validation} & \textbf{Generation} & \textbf{Validation} & \textbf{Generation} \\
\midrule
ViT-B/32 (dot prod.)       & $7.3 \pm 1.4$  & $5.3 \pm 2.4$  & $5.4 \pm 0.8$ & $3.5 \pm 1.9$ \\
ViT-B/32 (cosine)          & $8.9 \pm 1.3$  & $8.8 \pm 2.2$  & $6.7 \pm 0.7$ & $5.5 \pm 1.8$ \\
\addlinespace
ViT-L/14@336px (dot prod.) & $9.2 \pm 1.1$  & $6.2 \pm 2.7$  & $5.9 \pm 1.0$ & $4.9 \pm 2.0$ \\
ViT-L/14@336px (cosine)    & $9.8 \pm 1.0$  & $7.7 \pm 2.6$  & $7.1 \pm 1.0$ & $6.0 \pm 1.9$ \\
\addlinespace
RN50x64 (dot prod.)        & $9.8 \pm 1.2$  & $7.7 \pm 2.5$  & $6.4 \pm 0.9$ & $6.0 \pm 2.0$ \\
RN50x64 (cosine)           & $10.5 \pm 1.0$ & $8.2 \pm 2.6$  & $7.0 \pm 0.9$ & $5.5 \pm 2.0$ \\
\addlinespace
DINOv2 (dot prod.)         & $10.1 \pm 1.1$ & $10.2 \pm 2.6$ & $7.4 \pm 0.8$ & $8.1 \pm 2.1$ \\
DINOv2 (cosine)            & $10.0 \pm 1.0$ & $10.1 \pm 2.7$ & $7.3 \pm 0.8$ & $7.9 \pm 2.0$ \\
\addlinespace
ViT-g-14 (dot prod.)       & $11.9 \pm 0.9$ & $10.9 \pm 2.3$ & $9.0 \pm 0.8$ & $8.6 \pm 2.0$ \\
\textbf{ViT-g-14 (cosine)} & $\mathbf{12.3 \pm 1.0}$ & $\mathbf{11.1 \pm 2.3}$ & $\mathbf{9.3 \pm 0.8}$ & $\mathbf{8.6 \pm 1.9}$ \\
\bottomrule
\end{tabular}%
}
\end{table*}

\paragraph{Mathematical Sharpening and Fusion Strategy.}: Table \ref{tab:ablation_study} provides a detailed breakdown of our framework's components, where our tests highlight two critical findings. First, we observe that non-linear scaling is essential for improving the attribution results. By raising the similarity score to a high power, such as $p(c|x_j)^{10}$, the framework effectively rewards a small group of training images that hold actual influence in generative models while pushing the scores of unrelated images toward zero. Second, our proposed Asymmetric Rank Boost consistently performs better than standard linear fusion methods like 50/50 or 70/30 weighting. By using the Domain KG as a guide to re-rank the top visual matches, we maintain a clear visual signal while leveraging historical evidence to resolve ties. This ensures the final attribution is both visually plausible and historically accurate, allowing our black-box system to match the performance of more complex, gradient-based methods.

\begin{table*}[t]
\centering
\caption{Ablation study of our proposed attribution framework. LDS scores are reported. Our results show that raising the similarity score to a power ($p^{5}$ and $p^{10}$) significantly improves accuracy by rewarding the most relevant images and driving unimportant scores toward zero. Furthermore, our Asymmetric Rank Boost outperforms standard fusion by using the Domain's ranking to guide the visual comparison, ensuring the final attribution is both visually and semantically correct.}
\label{tab:ablation_study}
\resizebox{\textwidth}{!}{
\begin{tabular}{ll cccc}
\toprule
\multirow{2}{*}{\textbf{Modality}} & \multirow{2}{*}{\textbf{Formulation}} & \multicolumn{2}{c}{\textbf{ArtBench-2}} & \multicolumn{2}{c}{\textbf{ArtBench-5}} \\
\cmidrule(lr){3-4} \cmidrule(lr){5-6}
 & & \textbf{Validation} & \textbf{Generation} & \textbf{Validation} & \textbf{Generation} \\
\midrule
\multirow{4}{*}{\textbf{Visual}} 
 & $p(c|x_j)$ & $12.30 \pm 1.00$ & $11.10 \pm 2.30$ & $9.30 \pm 0.75$ & $8.60 \pm 1.90$ \\
 & $p(c|x_j)^5$ & $21.60 \pm 0.85$ & $18.00 \pm 2.25$ & $21.00 \pm 0.45$ & $16.80 \pm 1.45$ \\
 & $p(c|x_j)^{10}$ & $27.00 \pm 0.70$ & $20.10 \pm 1.80$ & $27.00 \pm 0.30$ & $17.50 \pm 1.05$ \\
 & $p(c|x_j)^{10}p(x_j)$ & $28.50 \pm 0.60$ & $20.80 \pm 1.70$ & $28.10 \pm 0.30$ & $18.00 \pm 1.00$ \\
\midrule
\multirow{4}{*}{\textbf{Domain}} 
 & $p(c|x_j)$ & $7.80 \pm 1.00$ & $6.20 \pm 1.50$ & $6.10 \pm 0.85$ & $4.60 \pm 1.35$ \\
 & $p(c|x_j)^5$ & $16.80 \pm 0.65$ & $11.90 \pm 1.35$ & $15.30 \pm 0.65$ & $10.90 \pm 1.35$ \\
 & $p(c|x_j)^{10}$ & $19.70 \pm 0.60$ & $12.30 \pm 1.30$ & $19.30 \pm 0.55$ & $10.70 \pm 1.40$ \\
 & $p(c|x_j)^{10}p(x_j)$ & $20.40 \pm 0.65$ & $12.40 \pm 1.20$ & $19.80 \pm 0.50$ & $10.80 \pm 1.35$ \\
\midrule
\multirow{3}{*}{\textbf{Hybrid (50/50)}} 
 & $p(c|x_j)$ & $10.90 \pm 0.85$ & $9.30 \pm 2.00$ & $8.60 \pm 0.75$ & $7.10 \pm 1.25$ \\
 & $p(c|x_j)^5$ & $20.80 \pm 0.70$ & $16.00 \pm 1.80$ & $20.00 \pm 0.60$ & $15.20 \pm 1.25$ \\
 & $p(c|x_j)^{10}$ & $25.60 \pm 0.65$ & $17.20 \pm 1.70$ & $25.70 \pm 0.45$ & $15.20 \pm 1.15$ \\
\midrule
\multirow{4}{*}{\textbf{Hybrid (70/30)}} 
 & $p(c|x_j)$ & $12.10 \pm 0.85$ & $10.70 \pm 2.15$ & $9.40 \pm 0.75$ & $8.30 \pm 1.35$ \\
 & $p(c|x_j)^5$ & $21.70 \pm 0.75$ & $17.30 \pm 2.00$ & $21.00 \pm 0.50$ & $16.30 \pm 1.30$ \\
 & $p(c|x_j)^{10}$ & $26.80 \pm 0.75$ & $18.90 \pm 1.85$ & $27.20 \pm 0.40$ & $16.70 \pm 1.10$ \\
 & $p(c|x_j)^{10}p(x_j)$ & $28.00 \pm 0.70$ & $19.40 \pm 1.55$ & $26.40 \pm 0.45$ & $15.40 \pm 1.10$ \\
\midrule
\textbf{Proposed} & \textbf{Asymmetric Rank Boost} & $\mathbf{29.10 \pm 0.65}$ & $\mathbf{21.30 \pm 1.65}$ & $\mathbf{28.90 \pm 0.35}$ & $\mathbf{18.40 \pm 1.05}$ \\
\bottomrule
\end{tabular}
}
\end{table*}

\section{Environmental KG: Schema and Construction}
\label{app:environment_kg}

To establish the necessary domain knowledge for the attribution in the environmental domain, we constructed a comprehensive KG ($\mathcal{G}_E$) that links physical flood events to their ecological and geographic surroundings. This allows the framework to evaluate weather predictions not just as pixel grids, but as events with real-world context.

\subsection{Data Collection and Source Integration}
To construct the environmental KG ($\mathcal{G}_E$), we developed a multi-source data pipeline anchored around specific historical weather events. Our foundational reference is based on a comprehensive registry of global extreme flood events \cite{extremekg}. This dataset provides the critical spatiotemporal anchors for our study, including unique event identifiers, the precise geographical coordinates of the region's event, temporal footprints, and macroscopic impact metrics such as human displacement and damage severity. 
While this foundational dataset establishes the macro-level context of extreme floods, evaluating and grounding predictions from generative meteorological models introduces a significant scale mismatch. Standard climate models typically generate predictions at a relatively coarse spatial resolution (e.g., a $0.25^\circ \times 0.25^\circ$ grid). At this macroscopic scale, it is impossible to directly map predictions to smaller, localised geographical features. To enable accurate, high-resolution data attribution, we must project these grid-level predictions into a richer semantic space. By augmenting the regional profiles with fine-grained external knowledge, we construct a domain representation that explicitly models the topological similarities, physical characteristics, and ecological interconnections between different disaster zones. This allows us to meaningfully compare historical events and trace potential lineages via retrieval at a much higher resolution than the models' native output grids.

To build this highly interconnected representation, we extended the geographical profiles of each flood region using an enrichment-based approach. We integrated the following high-resolution data sources:
\begin{itemize}
    \item \textbf{Biological Enrichment: iNaturalist\cite{inaturalist2026}.} Using the GPS anchors from the primary flood dataset, we enriched each event with localised ecological data. We queried the iNaturalist database to retrieve biological observations within a 10km radius of each flood center. This process provided verified taxonomic data (Species, Genus, and Kingdom), allowing the KG to encode the specific flora and fauna native to the region, which often serve as highly sensitive indicators of specific climate micro-zones and regional similarities.    
    \item \textbf{Geographic Enrichment: ESA LandCover\cite{zanaga2022esa}.} To capture the detailed physical characteristics of the disaster zones, we integrated the European Space Agency's (ESA) 10m-resolution LandCover data. By calculating zonal statistics for the 10km buffer around each epicenter, we extracted the physical characteristics of the land. This layer provides the percentage distribution of primary land-cover classes, such as tree cover, built-up urban areas, and mangroves.
\end{itemize}

\subsection{Ontological Schema and Neo4j Modeling}
All extracted data points were instantiated in a Neo4j graph database. By modeling these relationships structurally, we can ask complex questions about environmental vulnerability and historical context that raw, continuous meteorological data cannot answer. 

\begin{figure}[htbp]
    \centering
    \includegraphics[width=0.7\linewidth]{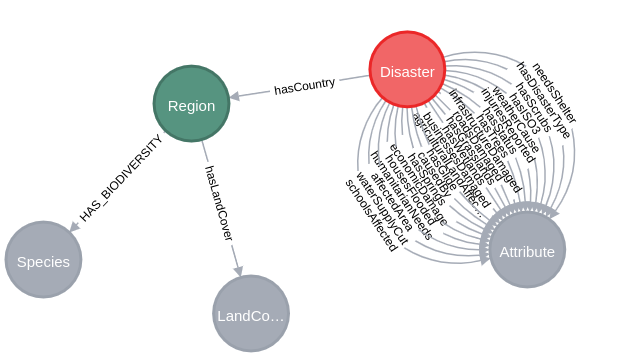}
    \caption{The ontological schema of the Environmental KG ($\mathcal{G}_E$). The schema defines the core nodes and directed edges used to map flood events through their regional hubs to their specific ecological and physical surroundings within Neo4j.}
    \label{fig:env_kg_schema}
\end{figure}

The ontological schema for the environmental domain centers on the \textbf{Disaster} event as the primary temporal anchor. This event is spatially linked to a \textbf{Region} node, which in turn acts as a node connecting to localised biological (\textbf{Species}) and geographic (\textbf{LandCoverClass}) nodes. The structural design of this ontology is illustrated in Figure \ref{fig:env_kg_schema}.

Table \ref{tab:env_kg_inventory} provides an inventory of the relationships currently within the graph, showcasing the semantic diversity of the dataset. This setup ensures that if a generative model predicts a flood in a specific geographic grid, our framework can retrieve historical events that share nearly identical environmental and topological contexts.

\begin{table}[ht]
\centering
\caption{KG Relationship Inventory and Semantic Diversity. We report the total edge count and unique target entities to demonstrate the breadth of the environmental context.}
\label{tab:env_kg_inventory}
\small
\begin{tabular}{llrrr}
\toprule
\textbf{Source} & \textbf{Relationship} & \textbf{Target} & \textbf{Count} & \textbf{Unique} \\
\midrule
Region   & Has\_Biodiversity & Species & 18,065 & 9,910 \\
Region   & hasLandCover      & LandCoverClass & 432 & 11 \\
Disaster & hasCountry        & Region & 1,592 & 149 \\
Disaster & hasDisasterType   & Attribute & 1,523 & 3 \\
Disaster & hasGlide          & Attribute & 1,468 & 1,468 \\
Disaster & causedBy          & Attribute & 1,197 & 471 \\
Disaster & weatherCause      & Attribute & 896 & 249 \\
Disaster & hasScrubs         & Attribute & 751 & 29 \\
Disaster & hasWetlands       & Attribute & 744 & 20 \\
Disaster & hasGrasslands     & Attribute & 703 & 22 \\
Disaster & affectedArea      & Attribute & 665 & 479 \\
Disaster & housesFlooded     & Attribute & 462 & 311 \\
Disaster & needsShelter      & Attribute & 407 & 11 \\
Disaster & economicDamage    & Attribute & 176 & 116 \\
\bottomrule
\end{tabular}
\end{table}

Table \ref{tab:env_node_edge_properties} details the internal properties and metadata fields associated with each node and edge type. These properties allow for precise filtering and the calculation of structural ranking scores within our pipeline.

\begin{table}[ht]
\centering
\caption{Entity and Relationship Property Schema of the Knowledge Graph $\mathcal{G}_E$.}
\label{tab:env_node_edge_properties}
\small
\begin{tabular}{ll}
\toprule
\textbf{Node Label} & \textbf{Internal Properties (Metadata Fields)} \\
\midrule
Disaster            & name, date, status, latitude, longitude, iso3, \\
                    & displacedPeople, housesCollapsed, powerOutage, \\
                    & waterSupplyCut, mountainsCount, watersCount, woodsCount \\
\addlinespace
Region              & name, latitude, longitude, springsCount, treesCount \\
\addlinespace
Species             & scientificName, category, inaturalistId, \\
                    & wikipediaUrl, photoUrl \\
\addlinespace
LandCoverClass      & name \\
\midrule
\textbf{Edge Type}  & Relationship Metrics (Edge Properties) \\
\midrule
hasLandCover      & area\_km2 \\
Has\_Biodiversity & observations \\
\bottomrule
\end{tabular}
\end{table}

Ultimately, this Neo4j structure allows for Direct Matching. Because each prediction has a location and time, we can query the graph to find the exact environmental footprint of the predicted event and compare it to historical records without the need for contrastive mapping.

\section{Environmental Evaluation Framework}
\label{app:weather_setup}
\subsection{Experimental Setup and Meteorological Data}
To evaluate our attribution framework on physical weather events, we leverage the pre-computed predictions of the Pangu-Weather model hosted by WeatherBench 2 \cite{rasp2023weatherbench}. Pangu-Weather optimises for both surface and upper-air variables by training on the ERA5 reanalysis dataset.

\paragraph{Temporal Scope and Forecast Intervals.}
Our evaluation leverages predictions from WeatherBench 2 dataset spanning a five-year period from 2018, to 2022. For each documented extreme flood event within this period, our primary experiments strictly utilise the 24-hour forecast target corresponding to the event's peak intensity. This specific lead time was chosen to effectively capture short-term, high-impact flood dynamics without the signal degradation.

\paragraph{Variables and Pressure Levels.}
The full Pangu-Weather output in WeatherBench 2 contains a comprehensive suite of atmospheric variables at a global $0.25^\circ \times 0.25^\circ$ spatial resolution. The available surface variables include 10m U and V wind components, 10m wind speed, 2m temperature, and mean sea-level pressure. The available upper-air variables include geopotential, specific humidity, temperature, U and V wind components, and wind speed. These upper-air variables are distributed across 13 distinct pressure levels ranging from 1000 hPa to 50 hPa.

\paragraph{Selected Features for Flood Analog Retrieval Case Study.}
To optimise the Latent Distribution Engine for our specific environmental case study, we subset this massive output tensor to isolate the physical drivers most relevant to extreme precipitation and surface flooding. We restrict the surface variables to 2m temperature and mean sea-level pressure. For the upper-air profile, we utilise temperature, specific humidity, geopotential, and the U/V wind components exclusively at the 1000, 850, 500, and 250 hPa levels. This targeted subset is physically justified: it captures essential low-level moisture transport and boundary layer dynamics (1000 and 850 hPa), mid-level steering flows (500 hPa), and upper-level divergence (250 hPa) necessary to attribute heavy rainfall events, effectively reducing computational overhead while preserving the primary meteorological signal.

\subsection{Multi-Modal Representation Learning}
\paragraph{Environmental Representation Learning.}
To bridge the gap between high-dimensional atmospheric tensors and discrete domain knowledge, we establish two specialised encoding pipelines. 

\textbf{Latent Representation (3D Denoising Autoencoder):} For the continuous meteorological modality, we developed a 3D ResNet-based Denoising Autoencoder. The encoder architecture utilises a series of 3D convolutional residual blocks to process the multi-level atmospheric cubes (surface and upper-air variables). We trained the model for 100 epochs on a dataset comprising ERA5 reanalysis data centered on extreme flood events and regional historical baselines. The resulting 512-dimensional embeddings provide a robust representation of the latent space for both ERA5 ground truth and Pangu-Weather predictions.

\textbf{Structural Representation (Node2Vec):} The environmental metadata is encoded via the Node2Vec algorithm \cite{node2vec} implemented within the Neo4j Graph Data Science (GDS) library. By utilising biased random walks over $\mathcal{G}_E$, we extract high-dimensional embeddings that capture the topological relationships between geographical regions, ecological biomes, and land-cover signatures. This ensures that the structural prior $P(\mathbf{X}_j)$ reflects the physical and ecological rarity of a region's position within the global environmental network, preventing common geographical hubs from biasing the analog retrieval.

\section{Computational Resources and Efficiency}
\label{app:compute}

To ensure the reproducibility of our results while maintaining a low computational footprint, we utilise a combination of pre-computed baseline assets and efficient graph-based infrastructure.

\subsection{Domain-Specific Assets}
\begin{itemize}
    \item \textbf{Artistic Domain:} We utilised pre-generated images and model weights provided by the D-TRAK baseline. This allowed us to evaluate attribution performance without the computational cost of retraining diffusion models from scratch and also ensure a fair comparison with baseline methods.
    \item \textbf{Environmental Domain:} We leveraged pre-computed Pangu-Weather predictions directly from the WeatherBench 2 benchmark, ensuring our physical-consistency evaluations are grounded in standardised meteorological AI outputs.
\end{itemize}

\subsection{Graph Infrastructure and Embedding Generation}
For the construction, hosting, and analysis of our domain-specific KGs, we utilised the Neo4j Desktop. A key advantage of this setup is that graph management and embedding generation did not require GPU access, relying instead on standard CPU compute. 

To generate embeddings natively based on our graph's topology and properties, we employed the Neo4j Graph Data Science (GDS) Library. The GDS library provides highly optimised algorithms that turn nodes and relationships into vector representations without leaving the database, allowing for efficient representation learning within the KG environment.

\subsection{Hardware and Implementation}
\begin{itemize}
    \item \textbf{Local Compute:} Local training for our contrastive MLPs and domain-specific encoders was performed on a single node equipped with one \textbf{NVIDIA RTX 5000 Ada Generation GPU (32GB VRAM)}.
    \item \textbf{Software Environment:} Our framework was implemented using PyTorch, the Diffusers library, and the D-TRAK requirements.
\end{itemize}

\section{Asset Licenses and Attribution}
\label{app:licenses}
We provide the licenses for all external assets used in this study in Table~\ref{tab:licenses}. 

\begin{table}[h]
\centering
\caption{Licenses and sources for external datasets and models.}
\label{tab:licenses}
\begin{tabular}{@{}lll@{}}
\toprule
\textbf{Asset} & \textbf{License} & \textbf{Sources} \\ \midrule
D-TRAK Codebase & MIT License & \cite{d-trak} \\
ArtBench Dataset & MIT License & \cite{artbench} \\
Environmental Data & MIT License & \cite{extremekg} \\
WeatherBench 2 & Apache 2.0 & \cite{rasp2023weatherbench} \\
Stable Diffusion & RAIL-M & \cite{hu2022lora} \\ \bottomrule
\end{tabular}
\end{table}


\end{document}